\documentclass[journal]{IEEEtran}
\ifCLASSINFOpdf
\else
\fi
\usepackage{cite}
\usepackage{amsmath,amssymb,amsfonts}
\usepackage{textcomp}
\usepackage{xcolor}

\usepackage{times}
\usepackage{soul}
\usepackage{url}
\usepackage[hidelinks]{hyperref}
\usepackage[utf8]{inputenc}
\usepackage[small]{caption}
\usepackage{graphicx}
\graphicspath{{Figures/}}
\usepackage{booktabs}

\usepackage[linesnumbered,lined,commentsnumbered, ruled]{algorithm2e} 

\usepackage{caption,subcaption} 
\usepackage{tabularx} 

\usepackage{tikz} \usetikzlibrary{positioning}
\usetikzlibrary{shapes,snakes}

\usepackage{balance}

\usepackage{multirow}

\newcommand{\BlackBox}{\rule{1.5ex}{1.5ex}}  

\newtheorem{theorem}{Theorem}

\newtheorem{definition}[theorem]{Definition}

\newcommand\independent{\protect\mathpalette{\protect\independenT}{\perp}}
\def\independenT#1#2{\mathrel{\rlap{$#1#2$}\mkern2mu{#1#2}}}

\begin{document}
%
\title{Deep Bayesian Estimation for Dynamic Treatment \\ Regimes  with a Long Follow-up Time}

%
%



\author{
   Adi~Lin, Jie~Lu, 
Junyu~Xuan, Fujin~Zhu,
  and~Guangquan~Zhang
  \thanks{A.~Lin, J.~Lu,J.~Xuan, F.~Zhu, G.~Zhang
    are with the Australian Artificial Intelligence Institute, FEIT, University
    of Technology Sydney, Australia (e-mail: Adi.Lin@student.uts.edu.au, Jie.Lu@uts.edu.au, Junyu.Xuan@uts.edu.au, Fujin.Zhu@uts.edu.au, Guangquan.Zhang@uts.edu.au).
  }
}

\maketitle

\begin{abstract}
Causal effect estimation for dynamic treatment regimes (DTRs) contributes to sequential decision making. However, censoring and time-dependent confounding under DTRs are challenging as the amount of  observational data declines over time due to a reducing sample size  but the feature dimension increases over time. Long-term follow-up compounds these challenges. Another challenge is the highly complex relationships between confounders, treatments, and outcomes, which causes the traditional and commonly used linear methods to fail. 
We combine outcome regression models with treatment models for high dimensional features using uncensored subjects that are  small in sample size and we fit deep Bayesian models for outcome regression models to reveal the complex relationships between confounders, treatments, and outcomes. 
Also, the developed deep Bayesian models can model uncertainty and output the prediction variance which is essential for the safety-aware applications, such as self-driving cars and medical treatment design. 
The experimental results on medical simulations of HIV treatment show the ability of the proposed method to obtain stable and accurate dynamic causal effect estimation from observational data, especially with long-term follow-up. Our technique provides practical guidance for sequential decision making, and policy-making.
\end{abstract}


%
\IEEEpeerreviewmaketitle













\section{Introduction}
\label{sec:intro}


\IEEEPARstart{M}{any} real-world situations require a series of decisions, so 
 a rule is needed (also referred to as a dynamic treatment regime (DTR), plan or policy) to assist making decisions at each step. For example, a doctor usually needs to design a series of treatments in order to cure patients. The treatment often varies over time, and the treatment assignment at each time point depends on the patient's history, which includes the patient's past treatments, the observed static or dynamic covariates that denote the patient’s characteristics, and the past measured outcomes of the patient's
disease. 
Estimating the causal effect of specific dynamic treatment regimes over time (also known as dynamic causal effect estimation) benefits decision makers when facing sequential decision-making problems. For example, estimating the effect of a series of treatments given by a doctor can apparently guide the real selection of correct treatments for patients. 
Although the gold standard for dynamic causal effect estimation is randomized controlled experiments, it is often either unethical, technically impossible, or too costly to implement \cite{spirtes2016causal}. For example, for patients diagnosed with the human immunodeficiency virus (HIV), it is unethical to pause their treatment for the sake of a controlled experiment. Hence, causal effect estimation is often conducted using uncontrolled observational data, which typically includes observed confounders, treatments, and outcomes. 

The challenges associated with estimating the causal effect of DTR are censoring and
time-dependent confounding. For example, the patients may quit the treatments
due to dead  over time and then the outcome values of these patients are
unknown (censoring). Furthermore, the features and outcomes of former time steps also
affect current treatments and current outcomes, so this increases the features
of current time point (time-dependent confounding). A longer follow-up makes
the challenges even more difficult. Another challenge is to address the highly complex
relationships between confounders, treatments, and outcomes, which causes the
traditional and commonly used linear methods to fail.


A motivating example is a HIV treatment research study with a focus on
children’s growth outcomes, measured by height-for-age z-scores (HAZ). The
subjects of the study are HIV-positive children who were diagnosed with AIDS,  before the commencement of the study. The treatment is antiretroviral (ART) which
is a dynamic binary treatment. The study is to evaluate the effect of
different treatment regimes applied to HAZ. The research took a couple of months
to observe the outcomes at a time point of interest. A child may be censored
during the study, where they may cease to follow the study during the treatment
process. Censoring leads to a  reduction in the number of subjects at
each time point in the HIV treatment research study, since not all the children may
survive the study after several months. ART may also bring drug-related
toxicities, so a judgement needs  to be made as to whether to use ART or not by the measurements
of a child's cluster of differentiation 4 count (CD4 count), CD4\%, and weight
for age z-score (WAZ). These measurements also serve as a disease progression
measure for the child. The CD4 count, CD4\%, and WAZ are affected by prior
treatments and also influence treatments (ART initiation) and the outcomes (HAZ) at
a later stage. Thus, CD4 count, CD4\%, and WAZ are time-dependent confounders.

Existing methods can be roughly categorised into three groups. The first group requires the treatment models  to fit for both the treatment and censoring mechanisms (these mechanisms model the treatment assignment process and the missing data mechanism), such as the inverse probability of treatment weighting (IPTW) \cite{austin2011introduction}. However, the cumulative inverse probability weights may be small due to the censoring problem, which leads to a near positivity violation (this violation makes it difficult to trust the estimates provided by the methods). The second group fits the outcome regression models (these models are to estimate the relationships between the outcome and the other variables), such as the sequential g-formula (seq-gformula) \cite{tran2019double}. However, the outcome regression models  may introduce bias into seq-gformula . 
The last group is a doubly robust method (LTMLE) \cite{schomaker2019using,van2018targeted}, which combines the iterated  
 outcome regressions and the inverse probability weights estimated by treatment models. However, it  suffers from the censoring problem because the observational data  reduces in the long-term which in turn decreases the modeling performance, especially with long-term follow-up.

In this paper, we propose a two-step model to combine the iterated outcome regressions
 and the inverse probability weights, where the uncensored subjects until the
 time point of interest are all used. 
 Hence our model improves the target
 parameter using potentially more subjects than LTMLE. LTMLE improves the target
 parameter using the uncensored subjects following the treatment regime of
 interest. It is for this reason that it is problematic to fit a limited number
 of subjects following the treatment regime of interest for common models in the
 long-term follow-up study. As our model uses a larger number of subjects in the
 estimation improvement, it may achieve better performance and stability. To capture the complex relationships between confounders, treatments,
 and outcomes in the DTRs, we apply a deep Bayesian method to the outcome
 regression models. The deep Bayesian method has a powerful capability to reveal complex hidden relationships using subjects of a small sample size. The
 experiments show that our method demonstrates the better performance and stability
 compared with other popular methods.

The three contributions of our study are summarised as follows:
\begin{itemize}
\item Our method is able to use all the uncensored data samples which cannot be fully used by other methods like LTMLE, so our method is able to achieve a more stable performance.
\item A deep Bayesian method is applied to the iterated outcome regression models.
  A deep Bayesian method has a strong ability to capture the complex hidden
  relationships between the confounders, the treatments, and the outcomes in a
  small-scale dataset with high dimensions.
\item The deep Bayesian method is able to capture the uncertainty for the causal
  prediction of every subject. Such uncertainty modeling is essential for the safety-aware applications, like self-driving cars and medical treatment design. 
\end{itemize}
  
The remainder of this paper is organized as follows.  Section~\ref{sec:related-work} discusses the related
work. Section~\ref{sec:preliminary-knowledge} describes the problem setting and a
deep Bayesian method.
We introduce our deep Bayesian estimation
for dynamic causal effect in Section~\ref{sec:our-proposed-model}.
Section~\ref{sec:experiments} describes the experiments with a detailed analysis of our method and its  performance compared with other methods.  Finally,
Section~\ref{sec:concl-future-work} concludes our study and discusses
future work.

\section{Related Work}
\label{sec:related-work}



In the majority of previous work,  learning the causal structure
\cite{huang2019bi,cui2017gaussian,xie2020entropy,yu2018markov,liu2018causal,cai2018merging} or conducting causal inference
from observational data in the static setting (a single time point)
\cite{yao2018representation,yao2019ace,tan2019user,kreif2019machine,hahn2020bayesian,lin2020causal} have been proposed. These methods cannot be applied to dynamic causal effect estimation 
\cite{bica2020counterfactual,hernan2020causal} directly. A few methods focus on the counterfactual prediction of outcomes for future treatments 
\cite{lim2018treatment,bica2020counterfactual,xu2016bayesian,schulam2017reliable}.
Counterfactual prediction methods aim to estimate the causal effect of the following future treatment, which is different from our problem setting. 

Another body of work focuses on selecting optimal DTRs \cite{zhang2019near,
  xu2016bayesian}. G-estimation \cite{murphy2003optimal, robins2004optimal} has
been proposed for optimal DTRs in the statistical and biomedical literature.
G-estimation builds a parametric or semi-parametric model to estimate the
expected outcome.
Two common machine
learning approaches, Q-learning \cite{liu2019learning} and A-learning
\cite{schulte2014q}, are applied to estimate DTRs. Q-learning is based on
regression models for the outcomes given in the patient information and is
implemented via a recursive fitting procedure. A-learning builds regret
functions that measure the loss incurred by not following the optimal DTR. It is
easy for the Q-functions of Q-learning and the regret functions of A-learning to
fit poorly in high dimensional data.

Finally, we discuss the methods proposed to estimate the dynamic causal
effect. The methods proposed to evaluate DTRs from observational data include
the inverse
probability of treatment weighting (IPTW) \cite{austin2011introduction}, the
marginal structural model (MSM) \cite{robins2000marginal,hernan2020causal}, 
sequential g-formula (seq-gformula) \cite{tran2019double}, and the doubly robust
method (LTMLE) \cite{schomaker2019using,van2018targeted}. 
IPTW estimates dynamic causal effect using the data from a
pseudo-population created by inverse probability weighting. The misspecified
parametric models used for the treatment models will lead to inaccurate estimation.
MSM fits a model that combines information from many treatment regimes to
estimate the dynamic causal effect. However, MSM may result in bias introduced by the regression model for the potential outcome and bias introduced by treatment models. Sequential g-formula uses the iterated conditional
expectation to estimate the average causal effect. The performance of sequential
g-formula relies on the accuracy of the outcome regression models. LTMLE calculates the
cumulative inverse probabilities by the treatment models on the uncensored
subjects who follow the treatment regime of interest. Then  it fits the outcome
regression models and uses the cumulative inverse probabilities calculated to
improve the target variable. The unstable weights from the treatment models may
affect the accuracy of LTMLE.

\section{Preliminary Knowledge}
\label{sec:preliminary-knowledge}

This section briefly introduces the problem setting using HIV treatment as an example and the deep kernel learning method.

\subsection{Dynamic Causal Effects}
\label{sec:dynam-caus-effects}

Observational data consists of information about the subjects. In the HIV
treatment study, patients are the subjects. For each subject, time-dependent
confounders $X_k$, treatment $T_k$ and outcome $Y_k$ are observed at
each time point $k$. Given the HIV treatment example, the treatment indicator is
whether ART is taken; CD4 variables are affected by previous treatments and
influence the latter treatment assignment and outcomes, CD4 variables belong to
time-dependent confounders; and HAZ is the outcome. The static confounders $V$
represent each subject's specific static features, such as patients' age. We use
$L_k$ to describe the union of static confounders $V$ and time-dependent
confounders $X_k$ at time $k$. 

The outcome values are unknown (censored) for
some subjects who pass away before follow-up. The censoring variable is represented as $C_k$.
If a subject does not follow the study, the subject will not attend the following study (if $C_i = 1$, then $C_j = 1$, for any time $j >
i$). If a subject follows the study at a time point, then the subject has
followed the study in all previous time points (if $C_i = 0$, then $C_j = 0$,
for any time $j < i$). Usually, a few subjects are censored at each time
point \cite{schomaker2019using}, and fewer subjects would continue to follow the study.

An example of censoring is given in the Table~\ref{tab:censoring}. At the
beginning, all subjects take part in the study. After the first
treatment assignment, some subjects do not follow the study. The features of
these subjects are ``missing'' in the following study. As the study continues,
an increasing number of subjects fail to follow the study. Thus, fewer subjects are
available in the following study, which becomes a
problem in the long follow-up study.

\begin{table*}[ht]
\caption{ 
An example of censoring. ``-'' represents a missing value.
}
\label{tab:censoring}
 \centering
\begin{tabular}{*{17}{c}}  
\toprule 
  Observation  &  $L_0$  & $C_0$ &  $Y_0$ & $T_0$ & $L_1$ & $C_1$ &  $Y_1$ & $T_1$ & $L_2$ & $C_2$ & $Y_2$ & \ldots & $T_K$ & $L_{K+1}$ & $C_{K+1}$ & $Y_{K+1}$ \\
\midrule
  $ob_0$ & $l_0^0$ & 0 & $y_0^0$ & $t_0^0$ & $l_1^0$ & 1 & - & - & - &- & - & \ldots & - & - &- &-\\
  $ob_1$ & $l_0^1$ & 0 & $y_0^1$ & $t_0^1$ & $l_1^1$ & 0 & $y_1^1$ & $t_1^1$ & $l_2^1$ & 1 & - & \ldots  & - & - &- &-\\
  $ob_2$ & $l_0^2$ & 0 & $y_0^2$ & $t_0^2$ & $l_1^2$ & 0 & $y_1^2$ & $t_1^2$ & $l_2^2$ & 0 & $y_2^2$ & \ldots  & $t_{K}^2$ & $l_{K+1}^2$ &0 & $y_{K+1}^2$ \\
\bottomrule
\end{tabular}
\end{table*}

To summarise, the observational data can be described as ($L_0$, $C_0$, $Y_0$),
($T_{k-1}$,$L_k$, $C_k$, $Y_k$), \text{ for } $k$ = 1, ..., $K$, and ($T_K$,
$L_{K+1}$, $C_{K+1}$, $Y_{K+1}$) in $K$+1 time point treatment regimes. All
subjects are uncensored at the baseline, which is $C_0 = 0$, and all subjects are
untreated before the study, which is represented as $T_{-1} = 0$. The $\bar{T}_k$ =
($T_0$, $T_1$, ..., $T_{k-1}$,$T_k$ ) represents the past treatments until the
time $k$. Other symbols with an overbar, such as $\bar{Y}_{k}, \bar{L}_{k}$, have
similar meanings. The history of covariates at time  $k$ is $H_k$ =
($\bar{T}_{k-1}$,$\bar{L}_{k}$,$\bar{C}_k$,$\bar{Y}_{k}$). For simplicity, the
subscript of variables will be omitted unless explicitly needed. Let
$Y^{\bar{d}}$ be the potential outcome of the possible treatment rule of
interest, $\bar{d}$, and $L^{\bar{d}}$ be the potential covariates of $\bar{d}$. The potential outcome
$Y^{\bar{d}}$ is the factual outcome or counterfactual value of the factual outcome.

 
 A treatment regime, also referred to as a strategy, plan, policy, or protocol,
 is a rule to assign treatment values at each time $k$ of the follow-up. A treatment
 regime is static if it does not depend on past history $h_k$. For example, the
 treatment regime ``always treat'' is represented by $\bar{d}_k = (1,1,...,1)$,
 and the treatment regime ``never treat'' is represented by $\bar{d}_k =
 (0,0,...,0)$. Both two treatment regimes assign a constant value for a
 treatment, so the assignments do not depend on past history. 
 A dynamic
 treatment regime depends on the past history $h_k$. The treatment assignment at the
 time point $k$ is $d_k = f(h_k)$. A dynamic treatment regime is a treatment
 rule $\bar{d}_k = (d_0,...,d_k)$. The outcome at the time point $k+1$ is affected
by the history $h_k$, the treatment $t_k$,
and the observed confounders $l_{k+1}$. The outcome can be described by $y_{k+1} = f(h_{k}, t_{k}, l_{k+1})$.

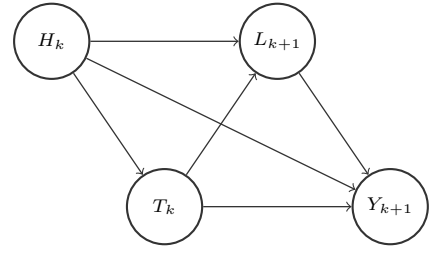
\begin{figure}[ht]
    \centering \tikzstyle{rv}=[circle, line width = 0.5pt, thick, minimum size= 1.0cm, 
    draw=black!80,
    ] \tikzstyle{drv}=[circle,line width = 0.5pt,  thick, dashed, minimum size= 1.1cm,
    draw=black!80,
    ] \tikzstyle{line}=[->, solid, line width=0.5pt, draw=black!80,
    ] \tikzstyle{dashline}=[->, dashed, line width=0.5pt, draw=black!80,
    ]

     \begin{tikzpicture}[font=\scriptsize,scale=1]
\node[rv] (1) at (0,2.2) {$H_{k}$};  
\node[rv] (2) at (1.5,0) {$T_{k}$};
\node[rv] (3) at (3,2.2) {$L_{k+1}$};  
\node[rv] (4) at (4.5,0) {$Y_{k+1}$};
     \path[line]  (1) edge (2)  (1) edge (3) (2) edge (3) ;
      \path[line]  (1) edge (4)  (2) edge (4) (3) edge (4) ;
     \end{tikzpicture}  
     
 \caption{The causal graphical model of time-dependent confounding for the
   time points $k$ and $k+1$. $H$ is the history of past treatments, confounders,
   and outcomes, \(T\) is the treatment, \(L\) is the observed
   confounders, and $Y$ is the outcome.}
  \label{fig:time-dependent-model}
\end{figure}

 The causal graph \cite{pearl:1995} for
dynamic treatments is illustrated in Fig.~\ref{fig:time-dependent-model}. The nodes represent the observed
variables. Links connecting the observed quantities are designated by arrows.
Links emanating from the observed variables that are causes to the observed variables
that are affected by causes. The treatment at 
time $k$ is influenced by the observed history $H_{k}$. The confounders $L_{k+1}$
and outcome $Y_{k+1}$ at  time $k+1$ are influenced by the history $H_{k}$.
The confounders $L_{k+1}$ also influence the outcome $Y_{k+1}$.

If all subjects are uncensored, the dynamic causal effect is to estimate the counterfactual mean outcome $E[Y^{\bar{d}_K}]$. To adjust the bias introduced by uncensoring in the data, the treatment assignment and uncensoring are considered as a joint treatment \cite{hernan2020causal}. The goal is to estimate the counterfactual mean
outcome $E[Y^{\bar{d}_K,\bar{c}_{K+1} = 0}]$. The mean outcome refers to the
mean outcome at time $K+1$ that would have been observed if all subjects have
received the intended treatment regime $\bar{d}$ and all subjects had been
followed-up.
The identifiability conditions for the dynamic causal effect need to
hold with the joint treatment ($T_m,C_{m+1}$) at all time $m$, where $m =
0,1,...,K$. Note that dynamic causal effect estimation from observational data is possible only with some
causal assumptions \cite{hernan2020causal}. Several common causal
assumptions are:
\begin{itemize}
    \item \textbf{Consistency}, that is $Y_{m}^{\bar{d}_{m}} = Y_{m}$ and
$\bar{L}_{m}^{\bar{d}_{m}} = \bar{L}_{m}$, if $\bar{T}_{m-1} = \bar{d}_{m-1}$.
The consistency assumption means the potential variable equals the observed
variable, if the actual treatment is provided.
\item \textbf{Positivity}, where $\Pr(T_{m} = \bar{d}_{m} | \bar{L}_{m} = \bar{l}_{m},\bar{Y}_{m} = \bar{y}_{m},
\bar{T}_{m-1} = \bar{d}_{m-1}) > 0$ for all $\bar{d}_{m}$, $\bar{l}_{m}$ and $\bar{y}_{m}$ with
$\Pr(\bar{T}_{m-1} = \bar{d}_{m-1}, \bar{L}_{m} = \bar{l}_{m}, \bar{Y}_{m} = \bar{y}_{m}) > 0$. The
positivity assumption means the subjects have a positive possibility of continuing
to receive treatments according to the treatment regime.
\item \textbf{Sequential ignorability}, it is $Y_{m}^{\bar{d}_{m}} \independent T_{m-1} |
\bar{L}_{m},\bar{Y}_{m-1}, \bar{T}_{m-2}$ for all $\bar{d}_{m}$, $\bar{l}_{m}$
and $\bar{y}_{m-1}$. The symbol $\independent$ represents statistical independence. The
sequential ignorability assumption means all confounders are measured in the data. 
\end{itemize} 
The causal assumptions consistency and sequential ignorability cannot be
statistically verified \cite{hernan2020causal}. A well-defined treatment regime
is a pre-requisite for the consistency assumption. The sequential ignorability
assumption requires that the researchers' expert knowledge is correct.

Based on these assumptions, the counterfactual mean outcome can be estimated from the observational data. Let $L_k$ absorb $Y_k$ if $k < K+1$ only in the Eq.~\eqref{eq:gformula} and Eq.~\eqref{eq:seqgformula}.
The counterfactual mean outcome
$E[Y^{\bar{d}_{K},\bar{c}_{K+1} = 0}]$ under the joint treatment ($\bar{d}_{K},\bar{c}_{K+1} =
0$) is identifiable through \textbf{g-formula}:
\begin{equation}
  \label{eq:gformula}
  \begin{aligned}[t]
    & \mathbb{E}[Y^{\bar{d}_{K},\bar{c}_{K+1} = 0}] =\\
    & \int_{\bar{l}_{K+1}}
      \mathbb{E}[Y_{K+1}|\bar{T}_{K} = \bar{d}_{K}, C_{K+1} = 0, \bar{L}_{K+1}=\bar{l}_{K+1}] 
      \\
     &\hspace{0.5cm} \cdot \prod_{k=0}^{K+1}f(l_k|\bar{T}_{k-1} = \bar{d}_{k-1},C_{k-1}=0, \bar{L}_{k-1} = \bar{l}_{k-1})
 \mathrm{d}\bar{l}_{K+1}, 
         \end{aligned}
  \end{equation}
with all the subjects remaining uncensored at time $K+1$. When $k=0$, $f(l_0)$ refers to the marginal distribution of $l_0$.
After integration with respect to $L$, the \textbf{iterated conditional
  expectation estimator} \cite{tran2019double} can be obtained using the iterative conditional expectation rule, where the equation holds that
\begin{equation}
  \label{eq:seqgformula}
  \begin{aligned}[t]
    & \mathbb{E}[Y^{\bar{d}_{K},\bar{c}_{K+1} = 0}] \\
    =& 
  \mathbb{E}[\mathbb{E}[\cdots \mathbb{E}[\mathbb{E}[Y_{K+1}|\bar{T}_{K}=\bar{d}_{K}, C_{K+1} = 
 0, \bar{L}_{K+1}] \\ & | \bar{T}_{K-1}
= \bar{d}_{K-1}, C_{K} = 0, \bar{L}_{K}]\cdots|\bar{T}_0 = d_0,C_1 = 0, L_1|L_0]].
  \end{aligned}
\end{equation}

\subsection{Deep Kernel Learning}
\label{sec:deep-kernel-learning}


Deep kernel learning (DKL) \cite{wilson2016deep} models are Gaussian process (GP)
\cite{rasmussen2006gaussian} models that use kernels parameterized by deep
neural networks. The definition and the properties of GPs are introduced first. Then, the
kernels parameterized by deep neural networks follow.

A Gaussian process can be used to describe a distribution over
functions with a continuous domain. The formal definition \cite{rasmussen2006gaussian} of
GPs is as follows,

\begin{definition}[Gaussian Process] \label{def:gp}
  A Gaussian process is a set of random variables, and any finite
  number of these random variables have a joint Gaussian distribution.
\end{definition}


A GP can be determined by its mean function and covariance function.
 The mean function
\(m(\mathbf{x})\) and the covariance function \(k(\mathbf{x},\mathbf{x^{'}})\)
of a GP \(g(\mathbf{x})\) are
\begin{align}
  m(\mathbf{x}) &= \mathbb{E}[g(\mathbf{x})], \label{eq:mean} \\
  k(\mathbf{x},\mathbf{x^{'}}) &= \mathbb{E}[(g(\mathbf{x})-m(\mathbf{x}))(g(\mathbf{x^{'}})-m(\mathbf{x^{'}}))]. \label{eq:kernel}
\end{align}
The features of the input are represented as $\mathbf{x}$.
The \(\mathbf{x}\) and \(\mathbf{x^{'}}\) are two input vectors.



Consider observations with
additive independent and identically distributed Gaussian noise, that is \(y = f(\mathbf{x}) + \varepsilon\). Observations can be represented as
\(\{(\mathbf{x_{i}},y_{i})| i = 1, \cdots, n\}\). The GP model for these observations is given by
\begin{equation}
  \label{eq:gp-noise}
  \begin{aligned}
    y(\mathbf{x}) &= f(\mathbf{x}) + \varepsilon, \varepsilon \sim \mathcal{N}(0, \sigma^2 ), \\
    f &\sim \mathcal{GP}(m,k), \\
    y &\sim \mathcal{GP}(m, k + \sigma^2 \delta_{i i^{'} }),
  \end{aligned}
\end{equation}
where $\delta_{i i^{'} }$ is the Kronecker's delta function, that is  $\delta_{i
  i^{'} } = 1$ if and only if $ i = i^{'} $.

The features to evaluate are denoted as \(\mathbf{x_{*}}\) and the outputs to
evaluate are represented as \(f_{*}\). The predictive distribution of the
GP---the conditional distribution \(f_{*}|X_{*},X,y\) is Gaussian, that is
\begin{equation}
  \label{eq:pred-gp}
  f_{*}|X_{*},X,y \sim \mathcal{N}(\mathbb{E}[f_{*} ], \operatorname{cov}(f_*)),
\end{equation}
where $\mathbb{E}[f_{*} ] = m(x_*) + K(X_*,X) [K(X,X) + \sigma^2 I ]^{-1}
(y - m(X)) $ and $\operatorname{cov}(f_*) = K(X_*, X) - K(X_*,X) [K(X,X) +
\sigma^2 I ]^{-1} K(X_*,X) $.

The log-likelihood function is used to derive the maximum likelihood estimator. The log likelihood of the outcome $\mathbf{y}$ is:
\begin{equation}
  \label{eq:log-gaussian}
  \log p(\mathbf{y}|\gamma, X) \propto - [\mathbf{y}^T (K_{\gamma} + \sigma^2 I)^{-1} \mathbf{y} + \log |K_{\gamma} + \sigma^2 I| ],
\end{equation}
where  $K_{\gamma}$ denotes kernel function $K(X,X)$
given the parameter $\gamma$.

DKL transforms the inputs of a base kernel with a deep neural network. That is,
given a base kernel $k(\mathbf{x}_i, \mathbf{x}_j|\mathbf{\theta})$ with hyperparameters
$\mathbf{\theta}$, the inputs $\mathbf{x}$ are transformed as a non--linear mapping $g(\mathbf{x},
\mathbf{w})$. The mapping $g(\mathbf{x},\mathbf{w})$ is given by a deep neural network
parameterized by weights $\mathbf{w}$.

\begin{equation}
  \label{eq:deep-kernel}
  k(\mathbf{x}_i, \mathbf{x}_j | \theta) \rightarrow k(g(\mathbf{x}_i,\mathbf{w}), g(\mathbf{x}_j,\mathbf{w})|\mathbf{\theta},\mathbf{w}).
\end{equation}

In order to obtain scalability, DKL uses the KISS-GP covariance matrix as the kernel
function $K_{\gamma}$,
\begin{equation}
  \label{eq:kiss}
  K_{\gamma} \approx M K_{U,U}^{deep} M^{T} := K_{KISS},
\end{equation}
where $ K_{U,U}^{deep}$  is a covariance matrix learned
by Eq.~\eqref{eq:deep-kernel},  evaluated over $m$ latent inducing points $U =
[\mathbf{u}_i]_{i=1 \ldots m}$, and $M$ is a sparse matrix of interpolation weights,
$M$ contains 4 non-zero entries per row for local
cubic interpolation.

The technique to train DKL is described briefly in the following content. DKL jointly
trains the deep kernel hyperparameters $\{w,\theta\}$, that is the weights of
the neural network, and the parameters of the base kernel. The training process
is to maximise the log marginal likelihood of the GP. The chain
rule is used to compute the derivatives of the log marginal likelihood with respect
to $\{w,\theta\}$. Inference follows a similar process of a GP.

Most deep learning models which cannot represent their uncertainty and usually perform poorly for small size observations.  Compared with most deep learning models, DKL may achieve good performance with  data of small size and DKL can capture the uncertainty for the predictions of the observations.

\section{Our proposed model}
\label{sec:our-proposed-model}

\begin{table}[ht]
  \caption{Key Notations.}
  \begin{minipage}{\columnwidth}
     \begin{center}
      \begin{tabular}{lp{0.75\columnwidth}}
         \toprule
         Symbol & Meaning \\
         \hline 
         \(T\) & treatment variable \\
         \(X\) & time-dependent confounders \\
         $V$ & static confounders \\
         $L$ & static confounders and time-dependent confounders \\
         \(Y\) & outcome variable\\
         $C$ & censoring variable \\
         $\bar{T}_k$ & the past treatments until the time point $k$ \\
         $\bar{L}_k$ & the past confounders until the time point $k$ \\
         $\bar{C}_k$ & the censoring history until the time point $k$ \\
         $\bar{Y}_k$ & the past outcomes  until the time point $k$ \\
         $\bar{d}$ & the treatment regime of interest \\
         $H$ & the treatments, the measured confounders and the outcomes history, the history until the time point $k$ is $ H_k =
(\bar{T}_{k-1},\bar{L}_{k},\bar{C}_k,\bar{Y}_{k})$ \\
         \(Y^{\bar{d}}\) & potential outcome of the treatment regime $\bar{d}$ \\
         \(Y^{\bar{d},\bar{c}}\) & potential outcome of the treatment regime $\bar{d}$ and censoring history $\bar{c}$  \\     
         \(\Pr()\) & probability function \\
         \bottomrule
      \end{tabular}
     \end{center}
  \end{minipage}
  \label{tab:notation}
 \end{table}

This section begins with an outline of our proposed method, followed by a
description of the methods applied in the treatment models and the outcome regression
models. 
Some key notations used throughout this paper are summarised in Table~\ref{tab:notation}.

\begin{table*}[ht]
    \caption{The baseline methods. IPTW and MSM only apply treatment models
      relating to both the treatment and censoring mechanism. Sequential
      g-formula only applies the outcome regression models  using the iterated
      conditional expectation estimator. Both LTMLE and our proposed method
      combine treatment models and outcome regression models to obtain a  dynamic
      causal effect.  }
    \label{tab:baselines}
 \begin{minipage}[t]{\linewidth}
    \centering
    \begin{tabular}{l|c|c|c|c}
      \toprule
  Method~~~~~~~~~~~~~~~~       &  ~~~~~~~~ID~~~~~~~~   & ~~~~~ Treatment models~~~~ &~~Outcome regression~~ & ~~Reference~~ \\
      \hline
      Inverse probability of treatment weighting & \textbf{IPTW}  & Yes & No & \cite{austin2011introduction}\\
      Marginal structural model & \textbf{MSM} & Yes & No & \cite{ hernan2020causal, robins2000marginal}\\
      Sequential g-formula & \textbf{Seq-} & No & Yes & \cite{tran2019double} \\
      Longitudinal targeted maximum likelihood estimation  & \textbf{LTMLE-}  &  Yes & Yes & \cite{schomaker2019using ,van2018targeted}\\
      Our proposed method & \textbf{TS-} & Yes  & Yes & \\
   \bottomrule
    \end{tabular}    
      \end{minipage}
    \end{table*}
 Firstly,
we aim to estimate the causal effect by using both the information from the outcome
regression models and the treatment models with as many subjects as possible. We
designed models for the treatment models on the uncensored subjects, which are
usually larger than the uncensored subjects following the treatment regime of
interest (these subjects are in LTMLE). Models may achieve better performance in data with larger observations.
Secondly, we develop
 a deep Bayesian method for the outcome regression models considering the 
 small-size but high dimensional data. The deep Bayesian method is believed to be superior in modeling this kind of data.

 As most approaches deal with censoring \cite{bang2005doubly,tran2019double,
schomaker2019using}, we consider the treatment and the
 uncensoring as a joint treatment. Suppose we are interested in the dynamic
 treatment regime $d$, our target is to estimate the counterfactual mean outcome
 $E[Y^{\bar{d}_{K},\bar{c}_{K+1}=0}]$. In the
 following context, we consider how our method can be implemented for a general situation with a binary
 treatment and a continuous outcome.

Since the treatment and the uncensoring are handled as a joint treatment, the
 treatment models relating to both the treatment and
censoring mechanism need to be estimated. The treatment model relating to the
treatment mechanism at the  time point $m$ is
\begin{equation}
  \label{eq:trm_tr}
  \Pr(T_m = 1|
        \bar{T}_{m-1},\bar{L}_{m},C_m = 0,\bar{Y}_m),
\end{equation}
which refers to the probability of uncensored subjects who are treated, after a $m-1$ time
points treated history.
The treatment model relating to the censoring mechanism
 at the  time point $m$ is
\begin{equation}
  \label{eq:trm_c}
  \Pr(C_{m+1}=0|\bar{T}_m,\bar{L}_{m+1},C_m =0, \bar{Y}_{m}),
\end{equation}
which is the probability of subjects who will follow the study at the time
point $m+1$.

For a dynamic treatment and time-dependent covariates, the inverse probability
weights need to be calculated at each time point. The inverse treatment
probabilities and inverse censoring probabilities are calculated separately.
The cumulative product of inverse treatment 
probabilities is calculated by
\begin{equation}
  \label{eq:cum_trm_tr}
   W^{\bar{T}_{m}} = \prod_{k=0}^{m}
          \frac{1}{\Pr(T_k|\bar{T}_{k-1},\bar{L}_k, C_k =0,
            \bar{Y}_{k})}.   \\
        \end{equation}
The cumulative product of inverse modified treatment 
probabilities is calculated by
\begin{equation}
  \label{eq:cum_trm_tr_md}
   W^{\bar{T}_{m}}_{md}  =  \frac{W^{\bar{T}_{m-1}}}{\Pr(T_m = d_m|\bar{T}_{m-1},\bar{L}_{m},C_m
            =0, \bar{Y}_{m})}.
        \end{equation}
The subjects have followed the causal study for the first $m-1$ time points.
The equation represents the probability of subjects taking a treatment $d_m$ at the time point
$m$.
        
 Similarly, the cumulative product of inverse  censoring
 probabilities is calculated by
 \begin{equation}
   \label{eq:cum_trm_c}
           W^{\bar{C}_{m+1} = 0} =  \prod_{k=0}^{m}
           \frac{1}{\Pr(C_{k+1}=0|\bar{T}_{k},\bar{L}_{k+1},C_k = 0,
            \bar{Y}_{k})}.   
 \end{equation}
 The cumulative product of inverse modified censoring
 probabilities is calculated by
 \begin{equation}
   \label{eq:cum_trm_c_md}
           W^{\bar{C}_{m+1} = 0}_{md} =  \frac{W^{\bar{C}_m =
              0}}{\Pr(C_{m+1}=0|\bar{T}_{m-1}, d_m,\bar{L}_{m+1},C_m =
            0,\bar{Y}_{m} )}.
 \end{equation}
 The equation represents the probability of subjects taking a treatment $d_m$ at the time point
$m$. These subjects have followed the causal study at the first $m-1$ time points.

The cumulative product of inverse treatment and censoring
probabilities is
\begin{equation}
  \label{eq:cum_trm_trc}
            W_{m} = W^{\bar{T}_{m}} \times W^{\bar{C}_{m+1} =
              0 }.  
\end{equation}
The cumulative product of inverse modified treatment and modified censoring
probabilities is
\begin{equation}
  \label{eq:cum_trm_trc_md}
              W_{m}^{md} = W^{\bar{T}_{m}}_{md} \times W^{\bar{C}_{m+1} =
                0 }_{md}.
            \end{equation}
The weights $W_{m}$ and $W_{m}^{md}$ are applied in the outcome regression models of our proposed method.            

\begin{figure}[!t]
  \centering \includegraphics[scale=0.76]{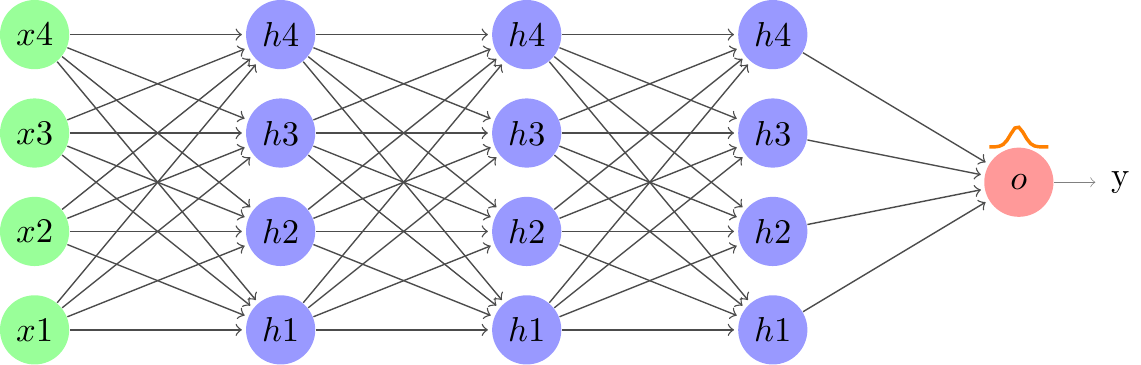}
 \caption{DKL for the outcome regression models.}
  \label{fig:dkl-out-reg-model}
\end{figure}

Another component of our proposed model is the outcome regression models.
Fit a DKL model for outcome regression models (Fig.~\ref{fig:dkl-out-reg-model}) at each time $m$, recursively from the last time point to the first time point.
The
outcome regression models are fitted and evaluated for all subjects that are
uncensored at each time point. 
At each iteration, DKL is trained for a outcome regression model firstly. Then DKL predicted an intermediate outcome used in the next iteration. The input is transformed with a deep neural network according to Eq.~\eqref{eq:deep-kernel}. At the iteration $m$, an outcome regression model
is fitted with the weight $W_m$ as a covariate.
 The input is the past treatments $\bar{T}_m$, the past observed confounders $\bar{L}_{m+1}$, the past outcomes $\bar{Y}_{m}$ and the weight $W_m$. 
And an intermediate outcome is predicted
by the estimated outcome regression model with the weight $W_{m}^{md}$ as a covariate. The intermediate outcome is used to train the DKL in the next iteration. 
The input to get the intermediate outcome is the treatments $\bar{T}_{m-1},d_m$, the past observed confounders $\bar{L}_{m+1}$, the past outcomes $\bar{Y}_{m}$ and the modified weight $W_{m}^{md}$. The output is a multivariate Gaussian distribution (Eq.~\eqref{eq:pred-gp}), which include a mean value (Eq.~\eqref{eq:mean}) and a variance value (Eq.~\eqref{eq:kernel}). The intermediate outcome is the mean value of the Gaussian distribution.


 \IncMargin{1em}
\begin{algorithm}[!t]
\caption{TS Algorithm}\label{alg:dr}
\SetKwData{Left}{left}\SetKwData{This}{this}\SetKwData{Up}{up}
\SetKwInOut{Input}{input}\SetKwInOut{Output}{output}
\Input{Treatments $T$, confounders $L$ and outcomes $Y$}
\Output{Causal effects}
\BlankLine
\For{$m\leftarrow 0$ \KwTo $K$}{

  Fit models for the treatment models relating to
  both the treatment and censoring mechanism (Eq.~\eqref{eq:trm_tr} and \eqref{eq:trm_c}).\\ 
Compute  $\hat{W}_{m}$  according to
Eq.~\eqref{eq:cum_trm_trc}  and  $\hat{W}^{md}_{m}$ according to
        Eq.~\eqref{eq:cum_trm_trc_md}. \\
        
                Set $\hat{Q}_{K+1}  = Y$. Then, for $m = K,K-1,\dots,0$, \\
      \qquad a) Fit a model $h(\bar{T}_m,\bar{L}_{m+1},\bar{Y}_{m}; \theta)$
        with a covariate $\hat{W}_{m}$ for the conditional expectation
        $\mathbb{E}[\hat{Q}_{m+1}|\bar{T}_{m},\bar{L}_{m+1},C_{m+1}=0,\bar{Y}_{m}]$
        with subjects that $C_{m+1} = 0$.   \\
        \qquad b) Predict the outcome $\hat{Q}_{m} $ based on the model $
        \hat{h}(\bar{T}_{m-1},d_m,\bar{L}_{m+1}, \bar{Y}_{m};\hat{\theta}) $ using the covariate $ \hat{W}^{md}_{m}$
        rather than  $\hat{W}_{m}$ with subjects that $C_{m} = 0$. \\
      The dynamic causal effect is calculated by the mean of the predicted outcome,
that is $\mathbb{E}[\hat{Q}_0]$. 
}
\end{algorithm}\DecMargin{1em}

The whole procedure of our algorithm is summarised in Algorithm~\ref{alg:dr}.

We use a logistic regression model on the assumption of a linear additive form of covariates to estimate time-varying
    inverse probability weights. Then, we apply DKL 
    \cite{wilson2016deep} for the outcome regression models.
    We name our two-step (TS) model
    with DKL for the outcome regression models, TS-DKL.

\section{Experiments}
\label{sec:experiments}


We design a series of experiments to evaluate the effectiveness of the proposed model. First, we introduce data simulations for dynamic treatment regimes. Then, we present the experimental setting and baseline methods, followed by an analysis of the experimental results. The experiments include the property analyses of the proposed model and the evaluation of the performance of the proposed model on simulated data.

\begin{figure*}[ht]
  \begin{equation}
    \label{eq:simdata}
    \small
    \begin{aligned}
    T_{-1} &\sim B(p=1) \\
    V^{1} &\sim B(p=4392 / 5826) \\
    V^{2} | \overline{\mathcal{D}}  &\sim\left\{\begin{array}{ll}{B(p=2222 / 4392)} & {\text {
                                                                      if } \quad
                                                                      V^{1}=1}
                                  \\ {B(p=758 / 1434)} & {\text { if } \quad
                                                         V^{1}=0}\end{array}\right.
                                                   \\
V^{3} |  \overline{\mathcal{D}} &\sim U(1,5)
\\
L_{0}^{1} | \overline{\mathcal{D}}
&\sim\left\{\begin{array}{ll}{N_{[0,10000]}(650,350)} & {\text { if } V^{1}=1} \\
             N_{[0,10000]}(720,400) & {\text { if }
                                                     V^{1}=0}\end{array}\right.
                                               \\
\tilde{L}_{0}^{1} | \overline{\mathcal{D}}  &\sim N\left(\left(L_{0}^{1}-671.7468\right) /(10
  \cdot 352.2788)+1,0\right) \\
 L_{0}^{2} | \overline{\mathcal{D}} &\sim N_{[0.06,0.8]}\left(0.16+0.05
   \cdot\left(L_{0}^{1}-650\right) / 650,0.07\right) \\
 \tilde{L}_{0}^{2} | \overline{\mathcal{D}} &\sim N\left(\left(L_{0}^{2}-0.1648594\right) /(10
   \cdot 0.06980332)+1,0\right) \\
L_{0}^{3} | \overline{\mathcal{D}} &\sim \left\{\begin{array}{ll} 
 N_{[-5,5]}( -1.65 + 0.1 \cdot V^{3}+0.05
   \cdot (L_{0}^{1}-650) / 650+0.05 \cdot(L_{0}^{2}-16) /
   16,1 )  &\text { if } V^{1} = 1 \\ 
N_{[-5,5]}(-2.05+0.1 \cdot V^{3}+0.05 \cdot(L_{0}^{1}-650) /
              650+0.05 \cdot(L_{0}^{2}-16) / 16,1) &\text{ if }
              V^{1} = 0
\end{array} \right. \\
C_{0} | \overline{\mathcal{D}} &\sim B(p=0) \\
Y_{0} |  \overline{\mathcal{D}} &\sim N_{[-5,5]} (-2.6 + 0.1 \cdot I(V^{3}>2) + 0.3 \cdot I(V^{1}
= 0) + (L_{0}^{3} + 1.45), 1.1) \\
T_{0} | \overline{\mathcal{D}} &\sim\left\{\begin{array}{ll}
 B(p=1) & \text { if } T_{-1}=1 \\
B\left(p=1 /\left(1+\exp \left(-\left[-2.4+0.015 \cdot\left(750-L_{0}^{1}\right)+5 \cdot\left(0.2-L_{0}^{2}\right)-0.8 \cdot L_{0}^{3}\right]\right)\right)\right) & \text { if } T_{-1}=0
                                           \end{array}\right.\\
\text{ For } &k > 0: \\
L_{k}^{1} |  \overline{\mathcal{D}} &\sim \left\{\begin{array}{ll}
N_{[0,10000]}\left(13 \cdot \log (k \cdot(1034-662) / 8)+L_{k-1}^{1}+2 \cdot L_{k-1}^{2}+2 \cdot L_{k-1}^{3}+2.5 \cdot T_{k-1}, 50\right) & \text { if } k \in\{1,2,3,4\} \\
N_{[0,10000]}\left(4 \cdot \log (k \cdot(1034-662) / 8)+L_{k-1}^{1}+2 \cdot L_{k-1}^{2}+2 \cdot L_{k-1}^{3}+2.5 \cdot T_{k-1}, 50\right) & {\text { if } k \in\{5,6,7,8\}} \\
N_{[0,10000]}\left(L_{k-1}^{1}+2 \cdot L_{k-1}^{2}+2 \cdot L_{k-1}^{3}+2 \cdot
                                    L_{k-1}^{3}+2.5 \cdot T_{k-1}, 50\right)  & \text { if } k \in\{9,10,11,12\}
\end{array}\right.\\
L_{k}^{2} | \overline{\mathcal{D}} &\sim N_{[0.06,0.8]}\left(L_{k-1}^{2}+0.0003 \cdot\left(L_{k}^{1}-L_{k-1}^{1}\right)+0.0005 \cdot\left(L_{k-1}^{3}\right)+0.0005 \cdot T_{k-1} \cdot \tilde{L}_{0}^{1}, 0.02\right)\\
L_{k}^{3} | \overline{\mathcal{D}} &\sim N_{[-5, 5]}\left(L_{k-1}^{3}+0.0017
  \cdot\left(L_{k}^{1}-L_{k-1}^{1}\right)+0.2
  \cdot\left(L_{k}^{2}-L_{k-1}^{2}\right)+0.005 \cdot T_{k-1}^{2} \cdot {\tilde{L}_{0}}^{2}, 0.5\right)\\
C_{k} | \overline{\mathcal{D}} &\sim B\left( max(p=1 /\left(1+\exp
    \left(-\left[-6+0.01 \cdot\left(750-L_{k}^{1}\right)+1
        \cdot\left(0.2-L_{k}^{2}\right)-0.65 \cdot
        L_{k}^{3}-T_{k-1}\right]\right)\right), 0.05) \right)\\
Y_{k} | \overline{\mathcal{D}} & \sim N_{[-5,5]}\left(Y_{k-1}+0.00005 \cdot\left(L_{k}^{1}-L_{k-1}^{1}\right)-0.000001 \cdot\left(\left(L_{k}^{1}-L_{k-1}^{1}\right) \cdot \sqrt{\tilde{L}_{0}^{1}}\right)^{2}+0.01 \cdot\left(L_{k}^{2}-L_{k-1}^{2}\right)-\right. \\
& 0.0001 \cdot\left(\left(L_{k}^{2}-L_{k-1}^{2}\right) \cdot
  \sqrt{\tilde{L}_{0}^{2}}\right)^{2}+0.07 \cdot\left(\left(L_{k}^{3}-L_{k-1}^{3}\right)
  \cdot\left(L_{0}^{3}+1.5135\right)\right)-0.001
\cdot\left(\left(L_{k}^{3}-L_{k-1}^{3}\right)
  \cdot\left(L_{0}^{3}+1.5135\right)\right)^{2} \\
& + \left.0.005 \cdot T_{k-1}+0.075 \cdot T_{k-2}+0.05 \cdot T_{k-1} \cdot T_{k-2},
2.5\right) \\
T_{k} | \overline{\mathcal{D}} &\sim\left\{\begin{array}{ll}
 B(p=1) & \text { if } T_{k-1}=1 \\
B\left(p=1 /\left(1+\exp \left(-\left[-2.4+0.015 \cdot\left(750-L_{k}^{1}\right)+5 \cdot\left(0.2-L_{k}^{2}\right)-0.8 \cdot L_{k}^{3}+0.8 \cdot k\right]\right)\right)\right) & \text { if } T_{k-1}=0
\end{array}\right.\\                                         
  \end{aligned}
\end{equation}
\end{figure*}

\subsection{Data simulation}
\label{sec:data-simulation}

Since rare ground truth is known in real non-experimental observational data, we
use simulation data from LTMLE \cite{schomaker2019using}
with a slight revision to fit the problem setting. We run ten simulations for the
data generation process. We simulate a binary treatment
indicator, a censoring indicator, three static covariates, three time-dependent
covariates and an outcome variable with twelve time points using causal
structural equations Eq.~\eqref{eq:simdata} by R-package
\emph{simcausal}~\cite{sofrygin2018simcausal}.

We simulate the static
covariates ($V^1, V^2, V^3$), which refers to  region, sex, age, respectively. The
time-dependent covariates ($L^1_{k}, L^2_{k}, L^3_{k}$) refer to CD4 count,
CD4\%, and WAZ at  time $k$, respectively. We simulate a binary treatment indicator $T_k$,
referring to whether ART was taken at time $k$ or not; a censoring indicator
$C_k$, describing whether the patient was censored (failing to follow the study)
at time $k$; and a continuous outcome $Y_k$, which refers to HAZ at time $k$. The binary
variable is simulated by a Bernoulli ($B$) distribution. We use uniform ($U$) distribution,
normal ($N$) distribution, and truncated normal distribution which is denoted by $N_{[a,b]}$
where $a$ and $b$ are the truncation levels, to simulate continuous variables.
When values are simulated in truncated normal distributions, if the values are
smaller than $a$,  they are replaced by a random draw from a $U(a_1, a_2)$
distribution. Conversely, if the values are greater than $b$, then the replacing values are
drawn from a $U(b_1, b_2)$ distribution. 
The values of ($a_1, a_2, b_1, b_2$) are (0, 50, 5000, 10000) for $L_1$,  
(0.03,0.09,0.7,0.8) for $L_2$, and $(-10, 3, 3, 10)$ for both $L_3$ and $Y$. The
notation $\bar{\mathcal{D}}$ denotes the data that have been observed before the
time point. All the subjects are untreated before the study and are represented by $T_{-1} =
0$, and all the subjects are uncensored at the baseline $k=0$.

The goal is to estimate the mean HAZ at time $K+1$ for the subjects. We consider four treatment regimes
(Eq.~\eqref{eq:tr_int}), where two treatment regimes are static, and the other two  
are dynamic.
\begin{equation}
  \label{eq:tr_int}
\begin{aligned}
&\bar{d}_{k}^{1}=\left\{c_{k}=0  \text { and }  t_{k}=1  \text { for all }   k\right.\\
&\bar{d}_{k}^{2}=\left\{\begin{array}{ll}
c_{k}=0  \text{ and } t_{k}=1 &\text { if condition 1}  \\
c_{k}=0  \text{ and } t_{k}=0 &\text { otherwise }
\end{array} \right. \\
&\bar{d}_{k}^{3}=\left\{\begin{array}{ll}
c_{k}= 0  \text{ and } t_{k}=1  &\text{ if condition 2}  \\
c_{k}= 0  \text{ and } t_{k}=0 &\text{ otherwise }
\end{array}  \right.\\
&\bar{d}_{k}^{4}=\left\{c_{k}= 0 \text { and }  t_{k}=0  \text { for all }  k\right. \\
& \text{condition 1:}\text { CD4 count }_{k}^{\bar{d}_{k}^{2}}<750  \text { or }  \mathrm{CD} 4 \%_{k}^{\bar{d}_{k}^{2}}<25 \%  \\ 
&  \text { or }  \mathrm{WAZ}_{k}^{\bar{d}_{k}^{2}}<-2  \text { or }  t_{k-1}=1 \\
& \text{condition 2:} \text { CD4 count }_{k}^{\bar{d}_{k}^{3}}<350  \text { or }  \mathrm{CD} 4 \%_{k}^{\bar{d}_{k}^{3}}<15 \%  \\ 
& \text { or }  \mathrm{WAZ}_{k}^{\bar{d}_{k}^{3}}<-2  \text { or }  t_{k-1}=1.
\end{aligned}
\end{equation}

The first and fourth treatment regimes are static. The first treatment regime
``always treat'' means all uncensored subjects are treated at each time point
during the study. The last treatment regime ``never treat'' means all uncensored
subjects are not treated at each time point during the study. The second and
third treatment regimes, ``750s'' and ``350s'', mean uncensored subjects receive
treatments until their CD4 reaches a particular threshold. Our proposed method is
 designed for the dynamic treatment regimes ``750s'' and ``350s''.

\subsection{Experimental setting}
\label{sec:experimental-setting}

The experimental setting is described briefly in this section.
The target is to evaluate the average dynamic causal effect, under a treatment
regime of interest, from the observational data. The evaluation metric for
estimation is the mean absolute error (MAE) between the estimated average causal
effect and the ground truth.

We use the Python package \emph{gpytorch} \cite{gardner2018gpytorch} to implement
the DKL model~\cite{liu2020deep,wilson2016deep}. We use a fully
connected network that has the architecture of three hidden layers with 1000,
500, and 50 units. The activation function is ReLU. The optimization method is
implemented in Adam \cite{kingma2014adam} with a learning rate of 0.01. We train the deep
kernel learning model for five iterations, and the DKL model uses
a GridInterpolationKernel (SKI) with an RBF base kernel.
A neural network feature extractor is used to
pre-process data, and the output features are scaled between 0 and 1.

\subsection{Baseline Methods}
\label{sec:baseline-methods}

We present the baseline methods which are the inverse probability of treatment weighting
(IPTW) \cite{austin2011introduction}, marginal structural model (MSM)
\cite{hernan2020causal, robins2000marginal}, sequential g-formula (Seq)
\cite{tran2019double}, and longitudinal targeted maximum likelihood estimation
(LTMLE) \cite{schomaker2019using ,van2018targeted}. The comparative methods and publication references are 
listed in Table~\ref{tab:baselines}. IPTW and MSM only apply treatment models,
sequential g-formula only applies outcome regression models, and LTMLE and our
proposed method apply both the treatment models and outcome regression models.
We describe the general implementation
procedure for these comparative methods as follows.

Given a fixed regime $\bar{d}$,  $\bar{g}_k(\bar{L}_{k+1}, \bar{Y}_{k})$ is
defined as Eq.~\eqref{eq:gfun},
\begin{equation}
 \label{eq:gfun}
 \footnotesize
 \begin{aligned}
 \bar{g}_{k}(\bar{L}_{k+1}, & \bar{Y}_{k}) = 
\prod_{m=0}^{k} \Pr(T_m = d_m|\bar{T}_{m-1} = \bar{d}_{m-1},C_{m} = 0, \bar{L}_m,\bar{Y}_m)  \\
 & \times \prod_{m=1}^{k+1} \Pr(C_m = 0 | \bar{T}_{m-1} = \bar{d}_{m-1}, C_{m-1}
    = 0, \bar{L}_m,\bar{Y}_{m-1}),
     \end{aligned}
   \end{equation}
where
$\bar{g}_{0}(\bar{L}_{1},Y_0) =
\Pr(T_0 = d_0|L_0,Y_0) \times  \Pr(C_1 = 0 | T_{0} = d_0, \bar{L}_1,Y_{0}).
$
 $\bar{g}_{n,k}(\bar{L}_{k+1}, \bar{Y}_{k})$ denotes the estimated values of
 $\bar{g}_{k}(\bar{L}_{k+1},\bar{Y}_{k})$.
 
Now we describe the general procedure of LTMLE \cite{schomaker2019using,van2018targeted}.
Before the outcome regression models with respect to the target variable are fitted to the observational data, we calculate the probabilities $\bar{g}_{n,k}(\bar{L}_{k+1}, \bar{Y}_{k})$ for each subject that
follows the given treatment regime $\bar{d}$. The following steps can be implemented as follows:
Set $\bar{Q}_{n,K+1} = Y$ (the continuous outcome should be rescaled to
  [0,1] using the true bounds and be truncated to (a,1-a), such as a = 0.0005).
  Then for $k = K,...,1$,
 \begin{enumerate}
\item
Fit a model for $\mathbb{E}[\bar{Q}_{n,k+1}|C_{k+1}=0, \bar{T}_{k},\bar{L}_{k+1}]$. The
model is fitted on all subjects that are uncensored until time $k+1$. \label{item:ltmle_step2.1}
\item
Plug in
$\bar{T}_{k} = \bar{d}_{k}$ based on the treatment regime $\bar{d}_k$, predict the
conditional outcome $\bar{Q}_{n,k}$ with the regression model from step~\ref{item:ltmle_step2.1} for all subjects with $C_{k}=0$. \label{item:ltmle_step2.2}
\item
\begin{itemize}
\item Plug in $\bar{T}_{k} = \bar{d}_{k}$ based on the treatment regime $\bar{d}_k$,
  get $\bar{Q}_{k} $ with the regression model from step~\ref{item:ltmle_step2.1} for all subjects with $C_{k+1}=0$. 
  \item  Construct the ``clever
    covariate''
    \begin{equation}
    \footnotesize
       CL_k(C_{k+1},\bar{T}_{k} = \bar{d}_{k},\bar{L}_{k+1},\bar{Y}_{k}) = \frac{ I(C_{k+1}=0,\bar{T}_{k} = 
   \bar{d}_{k})}{\bar{g}_{n,k}(\bar{L}_{k+1},\bar{Y}_{k})},
    \end{equation}

 where
   \(I(\cdot)\) is an indicator function for a logical statement. 
 \item
Run a no-intercept logistic regression. The outcome refers to
$\bar{Q}_{n,k+1}$, the offset is  $\text{logit}(\bar{Q}_{k})$, and  the covariate $CL_k$ is the unique
covariate. The model fits all subjects that are uncensored until time $k+1$ and
followed the treatment regime $\bar{T}_{k} = \bar{d}_{k}$.  Let $\hat{\epsilon}$ be
the estimated coefficient of $H_k$. 
\item
  Obtain the new predicted value of $\bar{Q}_{n,k}$ for all subjects with $C_{k}=0$
  by 
  \begin{equation}
      \bar{Q}_{n,k} = \text{expit}(\text{logit}(\bar{Q}_{n,k}) + \frac{\hat{\epsilon}_{k}}{\bar{g}_{n,k}(\bar{L}_{k+1},\bar{Y}_k)}).
  \end{equation}
\end{itemize}
\item
An estimate of the dynamic causal effect is obtained by taking the mean of
$\bar{Q}_{n,1}$ over all subjects; the mean is transformed back to the original
scale for the continuous
  outcome. \label{item:ltmle_step2.4}
\end{enumerate}

The sequential g-formula (the sequential g-computation estimator or the iterated
conditional expectation estimator) can be estimated by
step~\ref{item:ltmle_step2.1}, step~\ref{item:ltmle_step2.2}, and
step~\ref{item:ltmle_step2.4} of LTMLE.

IPTW uses the time-varying
    inverse probability weights to create pseudo-populations in order to estimate the dynamic
causal effect. The IPTW estimator for dynamic causal
effect at time $k$ can be obtained from $ \mathbb{E}[Y \cdot I(C_{k+1}=0,\bar{T}_{k} =
\bar{d}_{k})/\bar{g}_{n,k}(\bar{L}_{k+1},\bar{Y}_{k}) ] $, where $I(\cdot)$ is the indicator function.

To estimate the dynamic causal effect, we fit the ordinary
linear regression model $\mathbb{E}[Y^{\bar{d}_{k}, \bar{c}_{k+1}= 0}] =
\theta_{0} + \theta_{1} cum(\bar{d}_{k})$ to estimate the parameters of the
marginal structural model (MSM). The $cum(\bar{d}_{k})$ represents the cumulative
treatment in $k$ time points. MSM is fitted in the pseudo-population created by
inverse probability weights. That is, we use weighted least squares with inverse
probability weights.

To improve the performance of the outcome regression models,
Super Learner \cite{van2011targeted,van2007super} is often used. Super Learner is an ensemble machine learning method.
We test three different sets of learners for outcome regression models which are applied in
sequential g-formula, LTMLE, and our model. Learner 1 consists of ordinary
linear regression models, learner 2 contains ordinary linear regression models
and random regression forests, and learner 3 adds a multi-layer perceptron (MLP)
algorithm with a hidden layer with 128 units in addition to the algorithms used in learner 2. We
represent sequential g-formula with learners 1, 2 and 3 as Seq-L1, Seq-L2 and
Seq-L3, respectively. We represent similarly in LTMLE and our proposed method
with the three learners as the outcome regression models. In order to show the
effectiveness of  DKL, we also implement our proposed method
with a fully connected neural network (TS-NN) as the outcome regression model.
The network has the architecture of three hidden layers with 128, 64, and 32
units. The activation function is the ReLU activation function and dropout
regularization is used. We train our outcome regression model for 5 epochs with
a dropout rate of 0.9. The optimization method is implemented in Adam
\cite{kingma2014adam} with a learning rate of 0.01.

\subsection{ Propensity score truncation for positivity problems}
\label{sec:prop-score-trunc}
 
To calculate the dynamic causal effect, all baseline methods excluding
sequential g-formula have treatment models relating to both the treatment and
censoring mechanism. The treatment models for the $K+1$ time points is $2K+2$,
that is, $K+1$ treatment models relating to the treatment assignment mechanism, and
$K+1$ treatment models relating to the censoring mechanism. In long-term follow-up
causal studies, the treatment models need to regress on 
high-dimensional features using small-size samples at each time point. The
estimated  treatment and uncensored  probabilities may be small and the cumulative probabilities would be
close to zero. This may lead to near-positivity violations.

We deal with small treatment and uncensored probabilities by truncating them at a bound.
We truncate both the estimated treatment and uncensored probabilities at a lower
bound 0.1 and a higher bound 0.9. The truncation improves the stability of the
performance for the methods. In addition, we normalize the weights by their sum,
that is a new weight $W_i= W_i/ \sum_j W_j$. For all methods that apply the
treatment models, we use standard logistic regression models for the treatment
models relating to both the treatment and censoring mechanism.

\subsection{Evaluation of  a long follow-up time and declining sample size}
\label{sec:eval-long-foll}


Fewer subjects follow the study as the causal study progresses.
If a subject is uncensored at time $k$, then the subject has followed all the
studies in previous time points. If a subject is censored at time $k$, then the
variables related to the subject are unobserved from time $k+1$. The estimation of
dynamic causal effects is necessarily restricted to uncensored subjects. Another
challenge is that time-dependent confounders are affected by prior treatments
and influence future treatments and outcomes, hence it is necessary for the
time-dependent confounders to be adjusted. In long-term follow-up causal studies, 
high-dimensional adjustment confounders exist. All baseline methods are required
to fit the data of uncensored subjects at each time point. As a result, the estimators need to handle a reduced sample size and 
high-dimensional confounders at each time point.


\begin{table*}[ht]
\caption{ 
The number of features to regress increases, but the sample
size of the uncensored subjects to regress declines, over time. 
As the time-dependent confounding exists, more features need to regress but less
subjects are available in a longer study. The label ``uncensored'' refers to
the uncensored subjects. The labels ``750s'' and ``350s'' refer to
the uncensored subjects who have followed the dynamic treatment regime. 
}
\label{tab:highdim_anly}
 \centering
\begin{tabular}{*{13}{c}}  
\toprule 
~~~Time Point~~~  &  ~~~1~~~  & ~~~2~~~ &  ~~~3~~~ & ~~~4~~~ & ~~~5~~~ & ~~~6~~~ &  ~~~7~~~ & ~~~8~~~ & ~~~9~~~ & ~~~10~~~ & ~~~11~~~ & ~~~12~~~\\
\midrule
Dimension & 11 &  16 & 21 &  26  &31  &36 &41  &46  &51  &56  &61 &66 \\
  Uncensored & 883	&821	&777	&736	&699	&664 &632	&598	&571	&541	&514	&485\\
  750s & 480 & 429 & 397 & 374 & 353 & 333 & 315 & 297 & 283 & 266 & 250 & 235  \\
350s & 564 & 434 & 386 & 355 & 330 & 308 & 287 & 266 & 250 & 232 & 215 & 199  \\
\bottomrule
\end{tabular}
\end{table*}

We show that fewer subjects are available with more features to regress over time in Table~\ref{tab:highdim_anly}. There are 485 uncensored subjects
and 66 adjustment confounders at the twelfth time point, that is $K=11$. The
number of uncensored subjects at the time point of interest is quite small.
It is seen that the size of uncensored subjects is declining as the study
progresses (the time point increases) and it is noted that the number of features to
regress at each time point increases. Regression estimators need to fit data
with increasing features but smaller available observations.
Table~\ref{tab:highdim_anly} show that the number of uncensored subjects
following the dynamic treatment regime of interest  (``750s'' and ``350s'')
is even smaller than the number of uncensored subjects.

\begin{figure}[t]
  \centering \includegraphics[height = 4.5cm, width =
  \linewidth]{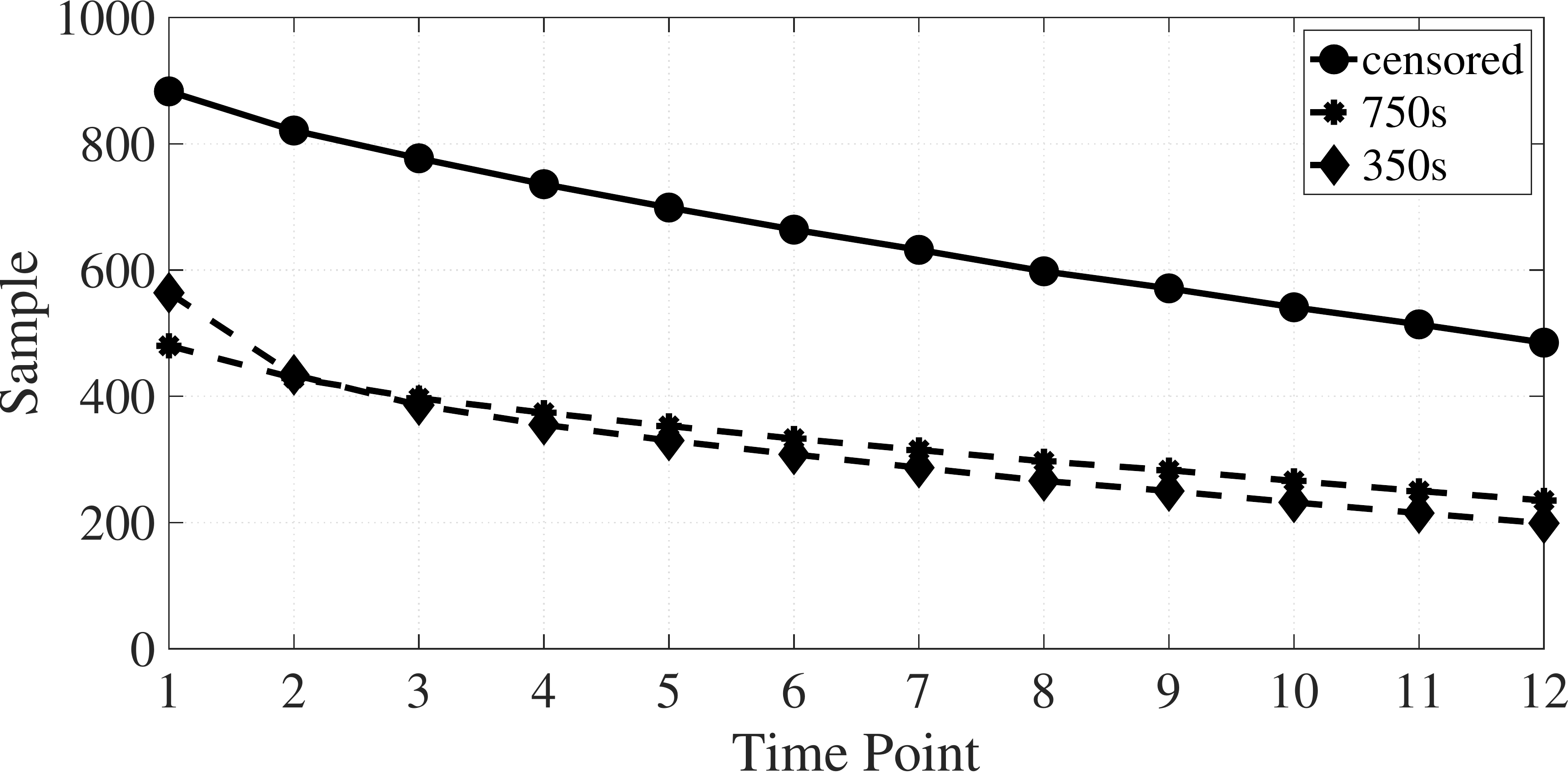}
  \caption{ The sample size of subjects declines as the study progresses. The
    label ``censored'' represents the number of subjects who are uncensored.
    The label ``750s'' represents the number of uncensored subjects who  
    receive the dynamic treatment regime ``750s''. The label ``350s'' represents
    the uncensored subjects who follow the treatment regime ``350s''. }
  \label{fig:tr_int_anly}
\end{figure}

The limited number of uncensored subjects who follow the treatment regime of
interest makes it challenging to estimate the dynamic causal
effect. Some estimators (such as LTMLE and IPTW) need to use the information from
the limited number of uncensored subjects who followed the treatment regime of
interest. However, these estimators may  have an unstable estimation in a long
follow-up study. We investigate the trend of two dynamic treatment regimes, ``750s'' and
``350s'', to show the declining, limited sample size of subjects who followed the
treatment regimes in Fig.~\ref{fig:tr_int_anly}. The size of the uncensored subjects
 reduces as the study progresses. This is also true for the uncensored
subjects following the treatment regime of interest, where
Fig.~\ref{fig:tr_int_anly} shows that the sample size of the uncensored subjects is
much larger than those uncensored subjects who followed a special treatment
regime. It is usually easier for a model to achieve stable performance by fitting the data with more
observations than fitting the data with less observations. The treatment models of our proposed
model fit the data of uncensored subjects. The treatment models of LTMLE fit the
data of uncensored subjects who followed the dynamic treatment regime. This
makes it possible for our proposed model to achieve a more stable estimation.

\subsection{Experimental analysis of our proposed method}
\label{sec:exper-perf-analys}

\begin{figure}[t]
  \centering \includegraphics[height = 4.5cm, width =
  \linewidth]{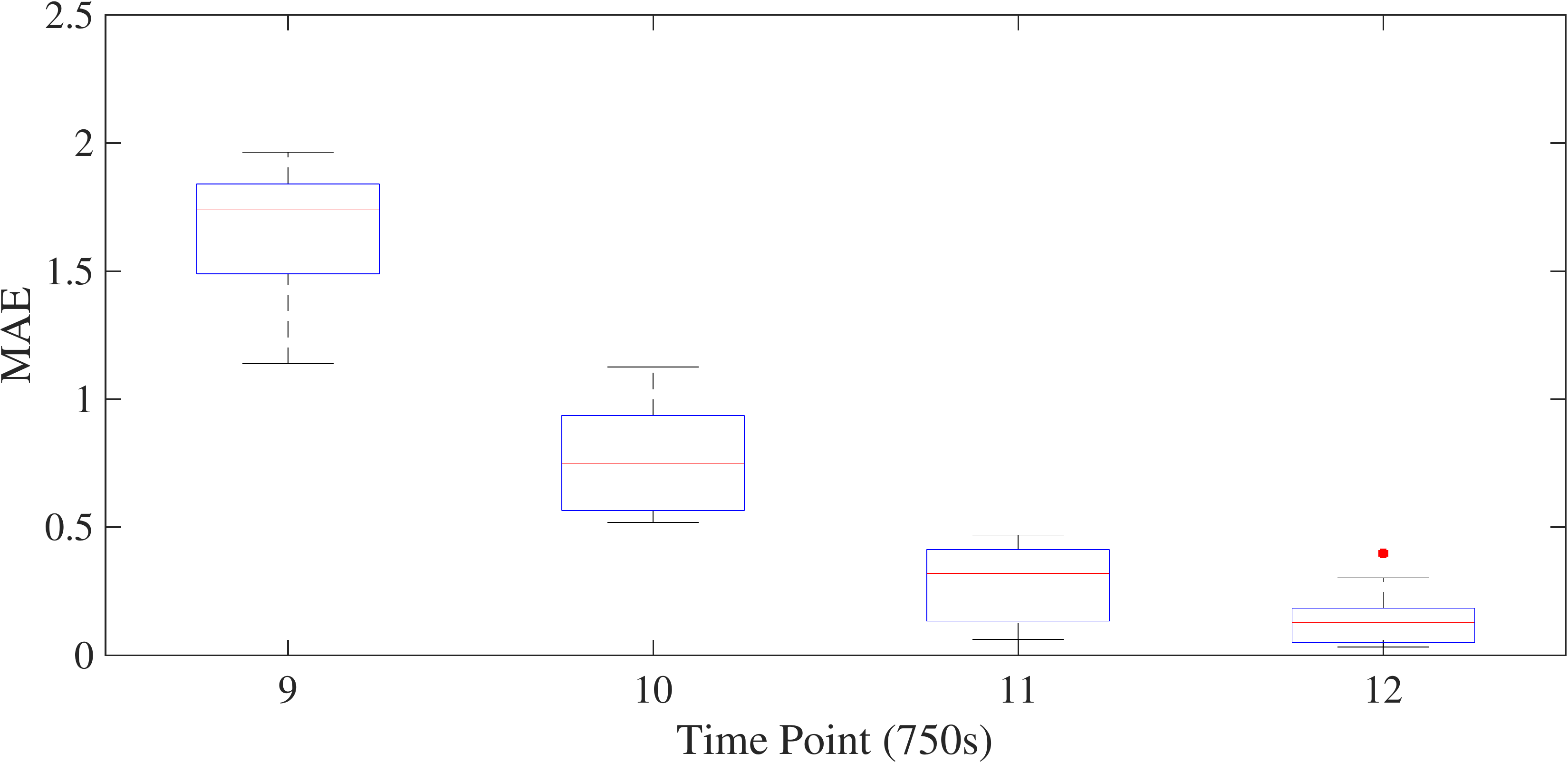}
  \caption{
    The experimental analysis for the performance (MAE) given the dynamic treatment regime,
    ``750s'', at different time $K+1=9, 10, 11, 12$. 
     }
  \label{fig:abar750s_long}
\end{figure}

\begin{figure}[t]
  \centering \includegraphics[height = 4.5cm, width =
  \linewidth]{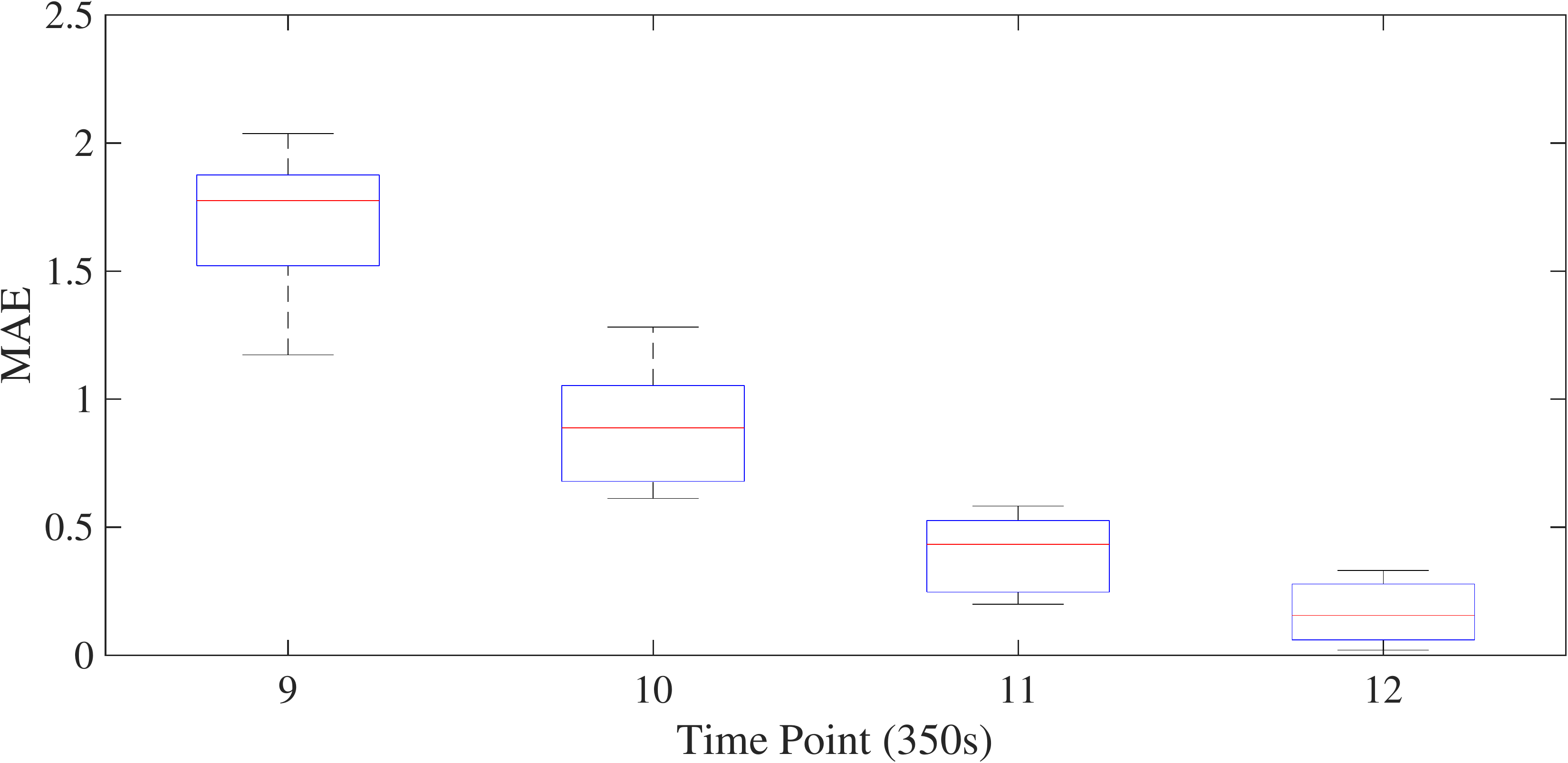}
  \caption{
    The experimental performance (MAE) of average causal effects for the treatment
    rule ``350s'' at different time points. We show the MAE for  time points
    9, 10, 11 and 12 ($K = 8, 9, 10, 11$).
     }
  \label{fig:abar350s_long}
\end{figure}

We analyse 
the influence of the number of subjects applied in the treatment models.
The treatment models of LTMLE and IPTW need to fit the limited number of
uncensored subjects who followed the treatment regime of interest at each time
point $k$. Unlike LTMLE and IPTW, the treatment models of our model fit on all
subjects who are uncensored at each time $k$. As more subjects are available in the 
treatment models, the estimate of our proposed method for a dynamic causal
effect may be more stable. DKL is applied in the outcome regression model at
each time point. DKL is suitable for the data which has high features but a  small number of 
observations. We show the MAE estimates for the two dynamic treatment regimes,
``750s'' and ``350s'', in Fig.~\ref{fig:abar750s_long} and
Fig.~\ref{fig:abar350s_long}. The MAE for the treatment regime, ``750s'', varies
between 0 and 2 for different time points. The MAE for the treatment regime,
``350s'', varies between 0 and about 2 for different time points. It is
interesting that bias reduces as the time point increases. Next, we focus on the
experimental analysis of the long follow-up, that is the time point 11 and 12.
As illustrated in Fig.~\ref{fig:abar750s_long} and Fig.~\ref{fig:abar350s_long},
the estimates of the dynamic causal effect with the given dynamic treatment regimes
(``750s'' and ``350s'') at different time points (the time point 11 and 12 ) do
not vary significantly. The figures show our model provides stable
estimates for the long-term follow-up causal study. It is also noted that the
bias of the dynamic causal effect at the time points 11 and 12 is small. These results indicate that our proposed method is able to achieve a stable and effective
estimate in a long-term follow-up study.

\begin{figure}[t]
  \centering \includegraphics[height = 4.5cm, width =
  \linewidth]{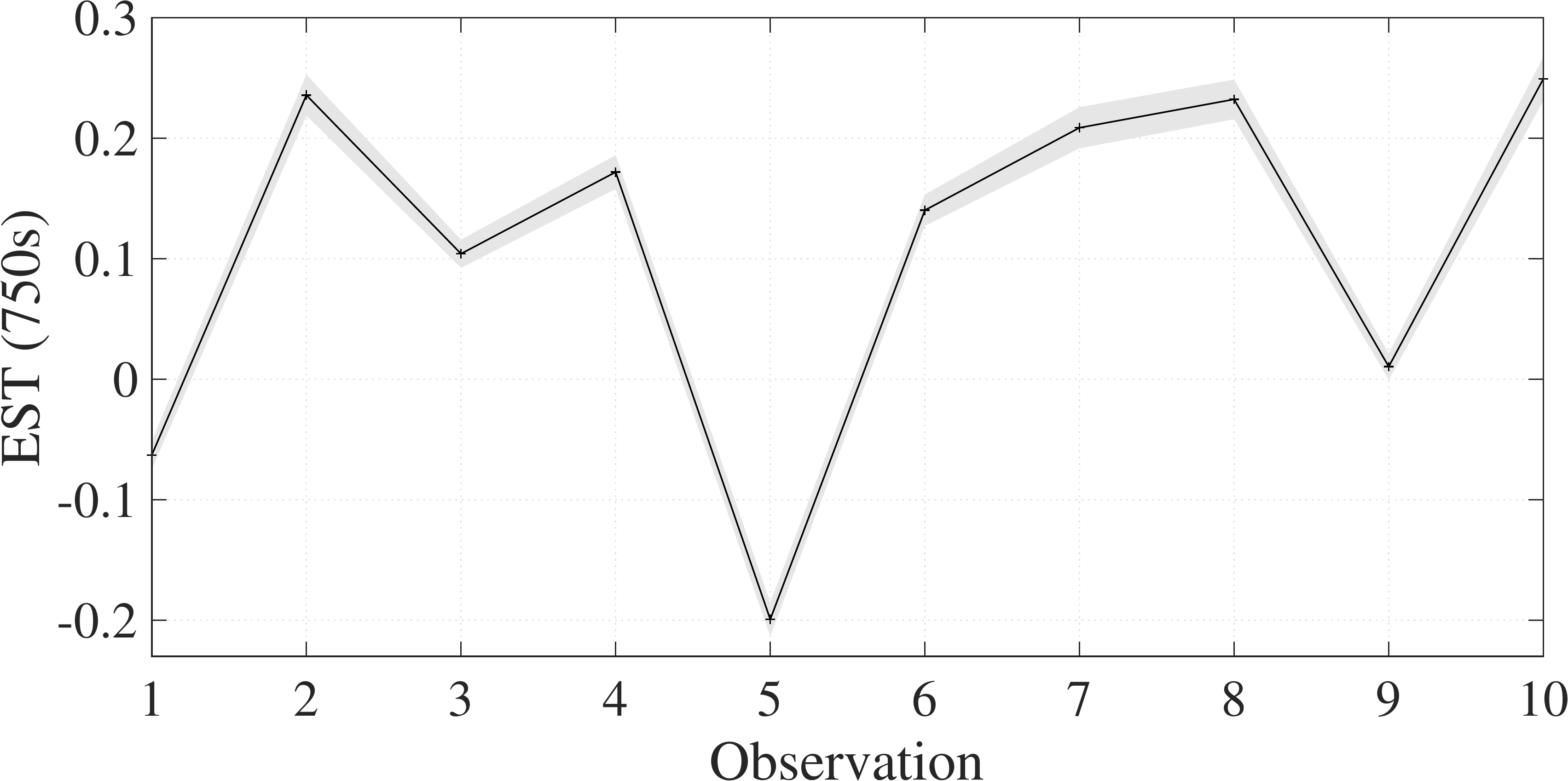}
  \caption{
    The plot shows the estimated value $\hat{y}^{\bar{d}_{K},\bar{c}_{K+1} = 
      0}$ and its variance for each observation. The number of the
    observations is 10. The time point is $K$+1 = 12.
     }
  \label{fig:abar750s_est_and_var}
\end{figure}

Next, we analyse DKL's ability to capture the variance for the causal prediction of a subject.
For every predicted value, DKL also produces its variance. Ten observations were
selected to show the predicted values $\hat{y}^{\bar{d}_{K},\bar{c}_{K+1} =
  0}$ and the variances of these predicted values.
Fig.~\ref{fig:abar750s_est_and_var} show the predicted values and its five
standard deviation. The variances are presented by the shade area in the figure.
It can be seen that the variance is small for each observation. DKL describes the
uncertainty of predictions through these small variances.

\subsection{Evaluation of dynamic causal effect with synthetic data}
\label{sec:eval-dynam-caus}


   
The results of the experiment using synthetic data are analysed in this section.   
We aim to estimate the dynamic causal effect $\mathbb{E}[Y^{\bar{d}_K,
  \bar{c}_{K+1} = 0}]$. The real dynamic causal effects are directly
obtainable within the simulated data. The performance is evaluated using the MAE
between the estimated causal effect and the real causal effect, for all ten
simulations. Our proposed method (TS-DKL) is designed for the dynamic treatment
regimes, ``750s'' and ``350s''. The comparative methods and publication
references are listed in Table~\ref{tab:baselines}. The outcome regression
models of sequential g-formula, LTMLE and our proposed method are equipped with
three different sets of learners, detailed in Section~\ref{sec:baseline-methods}. A
neural network is only implemented in the outcome regression model of our
proposed method (TS-NN). TS-NN is used to evaluate the effectiveness of the
DKL.

We summarise the experimental results of the estimates for the dynamic treatment
regimes, ``750s'' and ``350s'', in time points 11 and 12 in
Table~\ref{tab:sim_results_dynamic}. Table~\ref{tab:sim_results_dynamic} shows
that no models can obtain the best results for all estimates for the two dynamic
treatment regimes at the two different time points. We can observe that TS-DKL obtains the best
results (three treatments out of the four), followed by IPTW (one treatment out
of the four), in the two time points. TS-DKL achieves the second best
experimental performance for the dynamic treatment regime ``750s'' at the time
point 12, where the experimental performance of IPTW is the best. Compared with TS-NN, TS-DKL achieves better experimental performance in all dynamic treatment
regimes, hence we  conclude that applying DKL in the outcome
regression models is effective. The reason for this is the outcome regression model
needs to fit the data, which has high dimensions but is small in terms of  sample size. A deep
neural network usually works poorly with a limited number of observations. This is the
main reason that DKL is applied in the outcome regression models. Compared with the three TS-L* models, TS-L1, TS-L2, and TS-L3, we believe
DKL has a strong ability to capture the complex treatment,
time-varying confounders, and outcome relationships. The reason for this is that the only
difference between TS-DKL and the three TS-L* models is that the different methods are applied in the outcome
regression models. As the performance of  TS-DKL is the best
one in both two time points, TS-DKL has a strong ability to capture
the causal effects of dynamic treatment regimes in a long-term follow-up
causal study.

\begin{table}
\caption{ 
 The experimental results showing MAE from estimates of dynamic causal effect
 for  time
 points 11 and 12 ($K=10,11$). TS-DKL is our model with outcome regression models using deep kernel learning.
}
\label{tab:sim_results_dynamic}
\centering
\begin{tabular}{crrrr}  
  \toprule
  Time points & \multicolumn{2}{c}{11 } & \multicolumn{2}{c}{12 } \\
  \hline 
~~~Treatment~~~  &  750s & 350s & 750s & 350s \\
  \midrule
IPTW     & 0.2025 & 0.3042 & \textbf{0.0835} & 0.1932  \\
MSM      & 2.9740 & 3.0742 & 2.8150 & 3.0314  \\
Seq-L1   & 0.3043 & 0.1983 & 0.2997 & 0.1959  \\
Seq-L2   & 0.2363 & 0.1806 & 0.2742 & 0.1861  \\
Seq-L3   & 0.2536 & 0.2261 & 0.2739 & 0.2171  \\
LTMLE-L1 & 0.1744 & 0.1936 & 0.1863 & 0.1927  \\
LTMLE-L2 & 0.1475 & 0.1744 & 0.1658 & 0.1805  \\
LTMLE-L3 & 0.1324 & 0.1801 & 0.1724 & 0.1935  \\
\midrule
TS-L1    & 0.3031 & 0.1940 & 0.2926 & 0.1996  \\
TS-L2    & 0.1905 & 0.1900 & 0.2215 & 0.1887  \\
TS-L3    & 0.2197 & 0.2022 & 0.2650 & 0.2094  \\
TS-NN       & 0.3203 & 0.3751 & 0.3286 & 0.2726  \\  
TS-DKL   & \textbf{0.1253} & \textbf{0.1450} & \textbf{0.1528} & \textbf{0.1650} \\
\bottomrule
\end{tabular}
\end{table}

We analyse the experiment results  for our proposed method for the static treatment regimes. 
TS-DKL is designed to estimate the dynamic causal effects
for the dynamic treatment regimes. It is still unknown whether our proposed method
is able to achieve a comparative experimental performance for static treatment
regimes or not. Table~\ref{tab:sim_results_static} show the experimental performance of
the baseline methods for the static treatment regimes, ``always treat'' and
``never treat''. TS-DKL obtains the best
experimental performance under the static treatment regime ``always treat'', and
Seq-L3 obtains the best experimental performance under the static treatment
regime ``never treat'', for time point 11. IPTW obtains the best
experimental performance under the static treatment regime ``always treat'', and
TS-L2 obtains the best experimental performance under the static treatment
regime ``never treat'', for time point 12. It is noted that the
experimental performance of LTMLE-L3 and TS-DKL is relatively better than the other
baseline methods, excluding IPTW, under the static treatment regime ``always
treat'', for time point 12. Thus, the experimental performance of  TS-DKL under the treatment regime ``always treat'' is relatively
better than the other methods. All the baseline methods obtain poor estimates under the
treatment regime ``never treat''. One potential reason for this is the limited number of
uncensored subjects who followed the treatment regime ``never treat'', which reduces the models' ability to capture the data generating
process. Moreover, the limited number of subjects particularly affect the models
IPTW, MSM and LTMLE.

\begin{table}
\caption{ 
Experimental analysis with MAE of the estimated causal effects (time points 11 and
12) for the static treatment regimes. The TS-DKL is our model with a
deep kernel learning as the outcome
regression model.
}
\label{tab:sim_results_static}
\centering
\begin{tabular}{crrrr}  
  \toprule
  Time points & \multicolumn{2}{c}{11 } & \multicolumn{2}{c}{12 } \\
  \hline 
~~~Treatment~~~  &  all & never & all & never \\
  \midrule
IPTW     & 0.1798 & 0.7186 & \textbf{0.0583} & 0.6255  \\
MSM      & 3.1081 & 4.3238 & 3.2230 & 5.4558  \\
Seq-L1   & 0.3323 & 0.3216 & 0.3270 & 0.3255  \\
Seq-L2   & 0.2468 & 0.3059 & 0.2935 & 0.3172  \\
Seq-L3   & 0.2848 & \textbf{0.2692} & 0.3109 & 0.2979  \\
LTMLE-L1 & 0.1795 & 0.5780 & 0.1908 & 0.5502  \\
LTMLE-L2 & 0.1444 & 0.5463 & 0.1696 & 0.5119  \\
LTMLE-L3 & 0.1445 & 0.5390 & \textbf{0.1509} & 0.5279  \\
\midrule
TS-L1    & 0.3301 & 0.3361 & 0.3175 & 0.3497  \\
TS-L2    & 0.2042 & 0.3489 & 0.2376 & \textbf{0.2538}  \\
TS-L3    & 0.2724 & 0.3268 & 0.2368 & \textbf{0.2664}  \\
TS-NN    & 0.3109 & 0.7006 & 0.3433 & 0.4067  \\
TS-DKL   & \textbf{0.1251} & 0.5150 & \textbf{0.1595} & 0.4792 \\
\bottomrule
\end{tabular}
\end{table}

We analyse the bias introduced by the outcome regression models. 
The outcome regression models are applied in sequential g-formula, LTMLE and
our proposed method.
The results of the
empirical standard deviations (ESD) of the estimates for dynamic causal effect
in time points 11 and 12 are shown in
Table~\ref{tab:sim_results_esd}. TS-DKL performs stably under both dynamic treatment regimes in
the two different time points. Seq-L2 obtains stable experimental performance
under the dynamic treatment regime ``350s'', and obtains relatively larger ESD
under the dynamic treatment regime ``750s''. TS-L1 and TS-L2 have stable
performance under the dynamic treatment regime, ``350s'', in time point 12.
We can conclude that DKL is an  effective method to implement in the
outcome regression models. By comparising of TS-NN and TS-DKL, it can be seen that
Bayesian deep learning introduces robustness in the small-sized data. The neural
network applied in TS-NN performs unstably for the small-sized observations.

\begin{table}
\caption{ 
 The experimental results show the ESD from the estimates of dynamic causal effect
 for time
 points 11 and 12 ($K=10,11$). TS-DKL is our model with DKL as the outcome regression model.
}
\label{tab:sim_results_esd}
\centering
\begin{tabular}{crrrr}  
  \toprule
  Time points & \multicolumn{2}{c}{11 } & \multicolumn{2}{c}{12 } \\
  \hline 
~~~Treatment~~~  &  750s & 350s & 750s & 350s \\
  \midrule
Seq-L1   & 0.2896 & 0.1649 & 0.3086 & 0.2112 \\
Seq-L2   & 0.2332 & \textbf{0.1420} & 0.2873 & \textbf{0.2043} \\
Seq-L3   & 0.2529 & 0.1756 & 0.2908 & 0.2323 \\
LTMLE-L1 & 0.1942 & 0.1878 & 0.2408 & 0.2128 \\
LTMLE-L2 & 0.1942 & 0.1878 & 0.2408 & 0.2128 \\
LTMLE-L3 & 0.1619 & 0.1610 & 0.2256 & 0.2136 \\
\midrule
TS-L1    & 0.2812 & 0.1588 & 0.3000 & \textbf{0.2084} \\
TS-L2    & 0.1955 & 0.1721 & 0.2456 & \textbf{0.2056} \\
TS-L3    & 0.2534 & 0.2278 & 0.2931 & 0.2343 \\
TS-NN    & 0.4914 & 0.4915 & 0.3370 & 0.3374 \\
TS-DKL   & \textbf{0.1464} & \textbf{0.1415} & \textbf{0.1829} & \textbf{0.2072} \\
\bottomrule
\end{tabular}
\end{table}

\section{Conclusion and Future Work}
\label{sec:concl-future-work}

We proposed a deep Bayesian estimation for dynamic treatment regimes with
long-term follow-up. Our two-step method combines the outcome regression models with
treatment models and improves the target quantity using the information of
inverse probability weights on uncensored subjects. Deep kernel learning is
applied in the outcome regression models to capture the complex relationships
between confounders, treatments, and outcomes. The experiments have verified
that our method generally achieves both good performance and stability.

Currently, our approach lacks an analysis of asymptotic properties. The work on the
analytic estimation of standard errors and confidence intervals is yet to be
undertaken. In future work, we will provide theoretical guarantees for standard
errors and confidence intervals.






\ifCLASSOPTIONcaptionsoff
  \newpage
\fi



%

\bibliographystyle{IEEEtran}
\bibliography{ref}

\begin{thebibliography}{10}
\providecommand{\url}[1]{#1}
\csname url@samestyle\endcsname
\providecommand{\newblock}{\relax}
\providecommand{\bibinfo}[2]{#2}
\providecommand{\BIBentrySTDinterwordspacing}{\spaceskip=0pt\relax}
\providecommand{\BIBentryALTinterwordstretchfactor}{4}
\providecommand{\BIBentryALTinterwordspacing}{\spaceskip=\fontdimen2\font plus
\BIBentryALTinterwordstretchfactor\fontdimen3\font minus
  \fontdimen4\font\relax}
\providecommand{\BIBforeignlanguage}[2]{{%
\expandafter\ifx\csname l@#1\endcsname\relax
\typeout{** WARNING: IEEEtran.bst: No hyphenation pattern has been}%
\typeout{** loaded for the language `#1'. Using the pattern for}%
\typeout{** the default language instead.}%
\else
\language=\csname l@#1\endcsname
\fi
#2}}
\providecommand{\BIBdecl}{\relax}
\BIBdecl

\bibitem{spirtes2016causal}
P.~Spirtes and K.~Zhang, ``Causal discovery and inference: concepts and recent
  methodological advances,'' in \emph{Applied informatics}, vol.~3,
  no.~1.\hskip 1em plus 0.5em minus 0.4em\relax SpringerOpen, 2016, p.~3.

\bibitem{austin2011introduction}
P.~C. Austin, ``An introduction to propensity score methods for reducing the
  effects of confounding in observational studies,'' \emph{Multivariate
  behavioral research}, vol.~46, no.~3, pp. 399--424, 2011.

\bibitem{tran2019double}
L.~Tran, C.~Yiannoutsos, K.~Wools-Kaloustian, A.~Siika, M.~Van Der~Laan, and
  M.~Petersen, ``Double robust efficient estimators of longitudinal treatment
  effects: comparative performance in simulations and a case study,'' \emph{The
  international journal of biostatistics}, vol.~15, no.~2, 2019.

\bibitem{schomaker2019using}
M.~Schomaker, M.~A. Luque-Fernandez, V.~Leroy, and M.-A. Davies, ``Using
  longitudinal targeted maximum likelihood estimation in complex settings with
  dynamic interventions,'' \emph{Statistics in medicine}, vol.~38, no.~24, pp.
  4888--4911, 2019.

\bibitem{van2018targeted}
M.~J. van~der Laan and S.~Rose, \emph{Targeted Learning in Data Science: Causal
  Inference for Complex Longitudinal Studies}.\hskip 1em plus 0.5em minus
  0.4em\relax Springer, 2018.

\bibitem{huang2019bi}
H.~Huang, C.~Xu, and S.~Yoo, ``Bi-directional causal graph learning through
  weight-sharing and low-rank neural network,'' in \emph{2019 IEEE
  International Conference on Data Mining (ICDM)}.\hskip 1em plus 0.5em minus
  0.4em\relax IEEE, 2019, pp. 319--328.

\bibitem{cui2017gaussian}
R.~Cui, P.~Groot, and T.~Heskes, ``Robust estimation of gaussian copula causal
  structure from mixed data with missing values,'' in \emph{2017 {IEEE}
  International Conference on Data Mining, {ICDM} 2017, New Orleans, LA, USA,
  November 18-21, 2017}, 2017, pp. 835--840.

\bibitem{xie2020entropy}
F.~Xie, R.~Cai, Y.~Zeng, J.~Gao, and Z.~Hao, ``An efficient entropy-based
  causal discovery method for linear structural equation models with {IID}
  noise variables,'' \emph{{IEEE} Trans. Neural Networks Learn. Syst.},
  vol.~31, no.~5, pp. 1667--1680, 2020.

\bibitem{yu2018markov}
K.~Yu, L.~Liu, J.~Li, and H.~Chen, ``Mining markov blankets without causal
  sufficiency,'' \emph{{IEEE} Trans. Neural Networks Learn. Syst.}, vol.~29,
  no.~12, pp. 6333--6347, 2018.

\bibitem{liu2018causal}
F.~Liu and L.~Chan, ``Causal inference on multidimensional data using free
  probability theory,'' \emph{{IEEE} Trans. Neural Networks Learn. Syst.},
  vol.~29, no.~7, pp. 3188--3198, 2018.

\bibitem{cai2018merging}
R.~Cai, Z.~Zhang, Z.~Hao, and M.~Winslett, ``Sophisticated merging over random
  partitions: {A} scalable and robust causal discovery approach,'' \emph{{IEEE}
  Trans. Neural Networks Learn. Syst.}, vol.~29, no.~8, pp. 3623--3635, 2018.

\bibitem{yao2018representation}
L.~Yao, S.~Li, Y.~Li, M.~Huai, J.~Gao, and A.~Zhang, ``Representation learning
  for treatment effect estimation from observational data,'' in
  \emph{Proceedings of the 32nd International Conference on Neural Information
  Processing Systems}.\hskip 1em plus 0.5em minus 0.4em\relax Curran Associates
  Inc., 2018, pp. 2638--2648.

\bibitem{yao2019ace}
L.~Yao, S.~Li, Y.~Li, and M.~Huai, ``Ace: Adaptively similarity-preserved
  representation learning for individual treatment effect estimation,'' in
  \emph{2019 IEEE International Conference on Data Mining (ICDM)}.\hskip 1em
  plus 0.5em minus 0.4em\relax IEEE, 2019, pp. 1432--1437.

\bibitem{tan2019user}
F.~Tan, Z.~Wei, A.~Pani, and Z.~Yan, ``User response driven content
  understanding with causal inference,'' in \emph{2019 IEEE International
  Conference on Data Mining (ICDM)}.\hskip 1em plus 0.5em minus 0.4em\relax
  IEEE, 2019, pp. 1324--1329.

\bibitem{kreif2019machine}
N.~Kreif and K.~DiazOrdaz, ``Machine learning in policy evaluation: New tools
  for causal inference,'' in \emph{Oxford Research Encyclopedia of Economics
  and Finance}, 2019.

\bibitem{hahn2020bayesian}
P.~R. Hahn, J.~S. Murray, C.~M. Carvalho \emph{et~al.}, ``Bayesian regression
  tree models for causal inference: regularization, confounding, and
  heterogeneous effects,'' \emph{Bayesian Analysis}, 2020.

\bibitem{lin2020causal}
A.~Lin, J.~Lu, J.~Xuan, F.~Zhu, and G.~Zhang, ``A causal dirichlet mixture
  model for causal inference from observational data,'' \emph{ACM Transactions
  on Intelligent Systems and Technology (TIST)}, vol.~11, no.~3, pp. 1--29,
  2020.

\bibitem{bica2020counterfactual}
I.~Bica, A.~M. Alaa, J.~Jordon, and M.~van~der Schaar, ``Estimating
  counterfactual treatment outcomes over time through adversarially balanced
  representations,'' in \emph{8th International Conference on Learning
  Representations, {ICLR} 2020, Addis Ababa, Ethiopia, April 26-30,
  2020}.\hskip 1em plus 0.5em minus 0.4em\relax OpenReview.net, 2020.

\bibitem{hernan2020causal}
M.~Hern{\'a}n and J.~Robins, \emph{Causal Inference: What If}.\hskip 1em plus
  0.5em minus 0.4em\relax Chapman \& Hall/CRC, 2020.

\bibitem{lim2018treatment}
B.~Lim, ``Forecasting treatment responses over time using recurrent marginal
  structural networks,'' in \emph{Advances in Neural Information Processing
  Systems 31}, S.~Bengio, H.~Wallach, H.~Larochelle, K.~Grauman,
  N.~Cesa-Bianchi, and R.~Garnett, Eds.\hskip 1em plus 0.5em minus 0.4em\relax
  Curran Associates, Inc., 2018, pp. 7494--7504.

\bibitem{xu2016bayesian}
Y.~Xu, Y.~Xu, and S.~Saria, ``A bayesian nonparametric approach for estimating
  individualized treatment-response curves,'' in \emph{Machine Learning for
  Healthcare Conference}, 2016, pp. 282--300.

\bibitem{schulam2017reliable}
P.~Schulam and S.~Saria, ``Reliable decision support using counterfactual
  models,'' in \emph{Advances in Neural Information Processing Systems}, 2017,
  pp. 1697--1708.

\bibitem{zhang2019near}
J.~Zhang and E.~Bareinboim, ``Near-optimal reinforcement learning in dynamic
  treatment regimes,'' in \emph{Advances in Neural Information Processing
  Systems}, 2019, pp. 13\,401--13\,411.

\bibitem{murphy2003optimal}
S.~A. Murphy, ``Optimal dynamic treatment regimes,'' \emph{Journal of the Royal
  Statistical Society: Series B (Statistical Methodology)}, vol.~65, no.~2, pp.
  331--355, 2003.

\bibitem{robins2004optimal}
J.~M. Robins, ``Optimal structural nested models for optimal sequential
  decisions,'' in \emph{Proceedings of the second seattle Symposium in
  Biostatistics}.\hskip 1em plus 0.5em minus 0.4em\relax Springer, 2004, pp.
  189--326.

\bibitem{liu2019learning}
N.~Liu, Y.~Liu, B.~Logan, Z.~Xu, J.~Tang, and Y.~Wang, ``Learning the dynamic
  treatment regimes from medical registry data through deep q-network,''
  \emph{Scientific reports}, vol.~9, no.~1, pp. 1--10, 2019.

\bibitem{schulte2014q}
P.~J. Schulte, A.~A. Tsiatis, E.~B. Laber, and M.~Davidian, ``Q-and a-learning
  methods for estimating optimal dynamic treatment regimes,'' \emph{Statistical
  science: a review journal of the Institute of Mathematical Statistics},
  vol.~29, no.~4, p. 640, 2014.

\bibitem{robins2000marginal}
J.~Robins, M.~Hern{\'a}n, and B.~Brumback, ``Marginal structural models and
  causal inference in epidemiology.'' \emph{Epidemiology (Cambridge, Mass.)},
  vol.~11, no.~5, pp. 550--560, 2000.

\bibitem{pearl:1995}
J.~Pearl, ``Causal diagrams for empirical research,'' \emph{Biometrika},
  vol.~82, no.~4, pp. 669--688, 1995.

\bibitem{wilson2016deep}
A.~G. Wilson, Z.~Hu, R.~Salakhutdinov, and E.~P. Xing, ``Deep kernel
  learning,'' in \emph{Artificial Intelligence and Statistics}, 2016, pp.
  370--378.

\bibitem{rasmussen2006gaussian}
C.~E. Rasmussen and C.~K. Williams, \emph{Gaussian processes for machine
  learning}.\hskip 1em plus 0.5em minus 0.4em\relax MIT press Cambridge, 2006,
  vol.~1.

\bibitem{bang2005doubly}
H.~Bang and J.~M. Robins, ``Doubly robust estimation in missing data and causal
  inference models,'' \emph{Biometrics}, vol.~61, no.~4, pp. 962--973, 2005.

\bibitem{sofrygin2018simcausal}
O.~Sofrygin, M.~J. van~der Laan, and R.~Neugebauer, ``$\{$simcausal$\}$:
  Simulating longitudinal data with causal inference applications. 2015,''
  \emph{R package version 0.5}, vol.~3, 2018.

\bibitem{gardner2018gpytorch}
J.~Gardner, G.~Pleiss, K.~Q. Weinberger, D.~Bindel, and A.~G. Wilson,
  ``Gpytorch: Blackbox matrix-matrix gaussian process inference with gpu
  acceleration,'' in \emph{Advances in Neural Information Processing Systems},
  2018, pp. 7576--7586.

\bibitem{liu2020deep}
\BIBentryALTinterwordspacing
F.~Liu, W.~Xu, J.~Lu, G.~Zhang, A.~Gretton, and D.~J. Sutherland, ``Learning
  deep kernels for non-parametric two-sample tests,'' \emph{CoRR}, vol.
  abs/2002.09116, 2020. [Online]. Available:
  \url{https://arxiv.org/abs/2002.09116}
\BIBentrySTDinterwordspacing

\bibitem{kingma2014adam}
D.~P. Kingma and J.~Ba, ``Adam: A method for stochastic optimization,''
  \emph{arXiv preprint arXiv:1412.6980}, 2014.

\bibitem{van2011targeted}
M.~J. Van~der Laan and S.~Rose, \emph{Targeted learning: causal inference for
  observational and experimental data}.\hskip 1em plus 0.5em minus 0.4em\relax
  Springer Science \& Business Media, 2011.

\bibitem{van2007super}
M.~J. Van~der Laan, E.~C. Polley, and A.~E. Hubbard, ``Super learner,''
  \emph{Statistical applications in genetics and molecular biology}, vol.~6,
  no.~1, 2007.

\end{thebibliography}

\balance

\end{document}